\documentclass[review]{article}

\usepackage{epsfig}
\usepackage{graphicx}
\usepackage{subcaption}

\usepackage{amsmath}
\usepackage{amssymb}
\usepackage{mathtools}

\usepackage{adjustbox}
\usepackage{hhline}
\usepackage{tabularx}
\usepackage{booktabs}
\usepackage{multirow}
\usepackage[table,xcdraw]{xcolor}

\usepackage{textcomp}
\usepackage{lipsum}
\usepackage{pdflscape}
\usepackage{enumitem}
\usepackage{afterpage}
\usepackage{caption}
\usepackage{xcolor} 

\usepackage[pagebackref,breaklinks,colorlinks]{hyperref}

\setlistdepth{20} 
\setlist[itemize,1]{label=\textbullet}
\setlist[itemize,2]{label=--}
\setlist[itemize,3]{label=*}
\setlist[itemize,4]{label=+}
\setlist[itemize,5]{label=$\circ$}
\setlist[itemize,6]{label=$\cdot$}
\setlist[itemize,7]{label=\tiny$\blacksquare$}

\makeatletter
\newcommand{\mathleft}{\@fleqntrue\@mathmargin0pt}
\newcommand{\mathcenter}{\@fleqnfalse}
\newcommand{\subsubsubsection}[1]{\paragraph{#1}\mbox{}\\}
\makeatother

\definecolor{orcidlogocol}{HTML}{A6CE39} 

\usepackage{xr}
\begin{document}
\date{}
\title{A Hierarchical Graph-based Approach for Recognition and Description Generation of Bimanual Actions in Videos}
\author{
    Fatemeh Ziaeetabar$^{1,2,}$\thanks{Corresponding author: \href{mailto:fziaeetabar@gwdg.de}{fziaeetabar@gwdg.de}}  \href{https://orcid.org/0000-0003-1159-3588} {\includegraphics[scale=0.03]{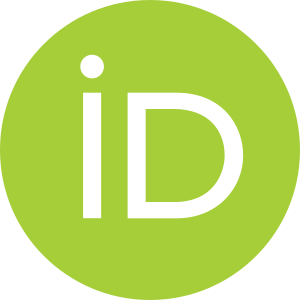}}, 
    Reza Safabakhsh$^{2}$\href{https://orcid.org/0000-0002-4937-8026}{\includegraphics[scale=0.03]{figures/id.png}}, 
    Saeedeh Momtazi$^{2}$\href{https://orcid.org/0000-0002-7271-9263}{\includegraphics[scale=0.03]{figures/id.png}},\\ 
    Minija Tamosiunaite$^{1,3}$\href{https://orcid.org/0000-0003-2996-3612}{\includegraphics[scale=0.03]{figures/id.png}},
    and Florentin W\"org\"otter$^{1}$\href{https://orcid.org/0000-0001-8206-9738}{\includegraphics[scale=0.03]{figures/id.png}}\\
    \small $^{1}$ Georg-Agust-Universit\"at G\"ottingen, G\"ottingen, Germany\\
    \small $^{2}$ Amirkabir University of Technology, Tehran, Iran\\
    \small $^{3}$ Vytautas Magnus University, Kaunas, Lithuania    
}
\maketitle
\begin{abstract}
Nuanced understanding and the generation of detailed descriptive content for (bimanual) manipulation actions in videos is important for disciplines such as robotics, human-computer interaction, and video content analysis. This study describes a novel method, integrating graph-based modeling with layered hierarchical attention mechanisms, resulting in higher precision and better comprehensiveness of video descriptions.

To achieve this, we encode, first,  the spatio-temporal interdependencies between objects and actions with scene graphs and we combine this, in a second step, with a novel 3-level architecture creating a hierarchical attention mechanism using Graph Attention Networks (GATs). The 3-level GAT architecture allows recognizing local, but also global contextual elements. This way several descriptions with different semantic complexity can be generated in parallel for the same video clip, enhancing the discriminative accuracy of action recognition and action description.

The performance of our approach is empirically tested using several 2D and 3D datasets. By comparing our method to the state of the art we consistently obtain better performance concerning accuracy, precision, and contextual relevance when evaluating action recognition as well as description generation. In a large set of ablation experiments we also assess the role of the different components of our model. With our multi-level approach the system obtains different semantic description depths, often observed in descriptions made by different people, too. Furthermore, better insight into bimanual hand-object interactions as achieved by our model may portend advancements in the field of robotics, enabling the emulation of intricate human actions with heightened precision.\\

    
\par{\textbf{Statements and Declarations:}}
The authors have no competing interests.\\

\end{abstract}



\section{Introduction}

Video understanding and description generation of manipulation and bimanual actions play a crucial role in various applications, including robotics, human-computer interaction, and video content analysis. The ability to accurately recognize, interpret, and describe human actions in videos is essential for enabling machines to understand and interact with the surrounding environment. Additionally, generating informative and contextually relevant video descriptions enhances user comprehension and facilitates efficient retrieval and analysis of video content.

Manipulation actions, involving the interaction between humans and objects, are fundamental to many tasks in robotics. They encompass a wide range of actions such as picking, grasping, manipulating, and releasing objects. Accurately recognizing and understanding manipulation actions is crucial for developing intelligent robotic systems capable of performing complex tasks in unstructured environments. By effectively analyzing manipulation actions, robots can adapt their behavior and manipulate objects with precision and dexterity, enhancing their utility in various domains, including manufacturing, healthcare, and home assistance.

Bimanual actions, which involve coordinated movements of both hands, are particularly important in robotics and human-robot collaboration scenarios. Bimanual actions enable humans and robots to perform tasks that require coordination, synchronization, and fine motor control. Examples include tasks such as assembling objects, manipulating tools, and operating complex machinery. By understanding and replicating bimanual actions, robots can better collaborate with humans, improving productivity, safety, and efficiency in shared workspaces.

Despite the significance of manipulation and bimanual actions, existing approaches to video understanding and description generation often face challenges in capturing the intricate relationships between objects and actions, modeling fine-grained spatial and temporal dependencies, and generating accurate and meaningful descriptions. These limitations hinder the development of robust and precise systems for manipulation and bimanual action recognition and description generation.

To address these challenges, we propose a novel methodology that combines graph-based modeling and hierarchical attention mechanisms. Our approach explicitly represents hand-object interactions as scene graphs, allowing for a comprehensive understanding of manipulation and bimanual actions by capturing the complex relationships between objects and actions. The utilization of graph-based modeling enables the modeling of fine-grained spatial and temporal dependencies, facilitating accurate recognition and interpretation of actions.

Furthermore, our methodology incorporates hierarchical attention mechanisms, specifically Graph Attention Networks (GATs), which capture local and global contextual information and enhance the precision and discriminative power of action recognition. By attending to relevant objects and temporal features, our model effectively aligns video content with corresponding textual descriptions, resulting in accurate and contextually relevant video descriptions.

To evaluate the effectiveness of our methodology, we conduct extensive experiments on diverse 2D and 3D datasets, encompassing a wide range of manipulation and bimanual actions. The experimental results demonstrate the superiority of our approach compared to state-of-the-art methods in terms of accuracy, precision, and contextual relevancy. Additionally, we introduce an enriched bimanual action taxonomy that improves the understanding and classification of actions, further enhancing the generation of accurate and nuanced video descriptions.


\begin{figure}[!htbp]
    \centering
   \includegraphics[scale=0.95]{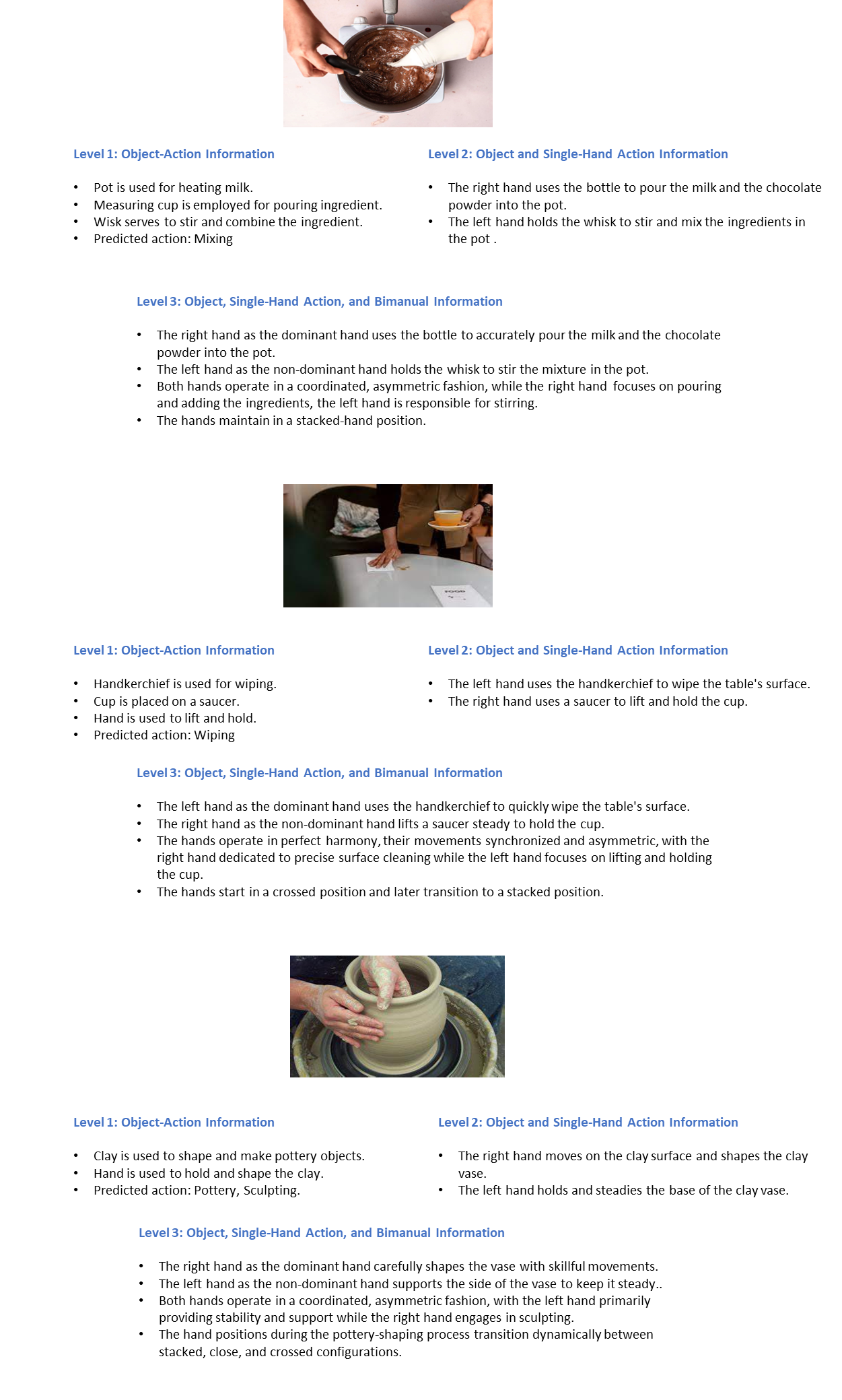}
    \caption{Hierarchical framework for generating video descriptions at different levels of detail. The figures include three video frames and examples of their generated descriptions at three levels of detail: object-level (Level 1), single hand action-level (Level 2), and bimanual action-level (Level 3).}
    \label{Fig.three_levels}
\end{figure}

The contributions of this paper can be summarized as follows: 

\begin{enumerate}
\item We propose a novel methodology that integrates graph-based modeling and hierarchical attention mechanisms to achieve highly accurate video understanding and description generation of manipulation and bimanual actions. To achieve this we use a novel ``layered Graph Attentional Network (GAT)" architecture. 


\item The layered architecture makes it possible to design a system that describes actions with different levels of detail, allowing us to flexibly generate concise summaries for quick overviews or elaborate and informative, paragraph-like captions that provide in-depth insights into the video content. This adaptability in capturing action complexity makes our model compatible with a wide range of applications, from video browsing and retrieval to educational content creation and video analysis in various domains. The concept of descriptions at different levels is depicted in Fig.~\ref{Fig.three_levels}, which had been created by our methods.

\item We conduct extensive experiments on diverse 2D and 3D datasets to evaluate the performance of our methodology. The evaluation includes metrics such as accuracy, precision, and contextual relevance. The results demonstrate the superiority of our approach compared to state-of-the-art methods, highlighting its effectiveness in accurately understanding and describing manipulation and bimanual actions.

\end{enumerate}

The remainder of this paper is organized as follows: Section \ref{litereture} provides a review of related work in video understanding, description generation, and manipulation and bimanual action recognition.

Section \ref{methods} presents the details of our proposed methodology. We describe the key components of our approach, including graph-based modeling, hierarchical attention mechanisms, and an enriched bimanual action taxonomy. We explain how these components work together to achieve accurate video understanding and description generation of manipulation and bimanual actions. Note that we have moved many implementation aspects, notably all learning processes, into the Supplementary Material, where algorithmic procedures are described in great detail.

Section \ref{results} is dedicated to the experimental results and analysis. We provide detailed explanations of the datasets used in our evaluation, and present the performance comparisons of our approach against state-of-the-art methods. In addition, we show results from a set of ablation experiments to demonstrate the contributions of the different algorithmic components. In this section, we also discuss the results, highlight the strengths of our approach, and provide insights into the potential applications of our methodology.

Finally, Section \ref{conclusion} concludes the paper by summarizing the contributions of our work and discussing future research directions.

\section{RELATED WORKS}
\label{litereture}

The domain of video understanding, encompassing tasks such as action recognition and video description generation, has experienced substantial progress in recent years. However, understanding and describing complex bimanual actions in videos remains a challenge due to the intricacies involved in capturing the relationships between objects and actions. This section provides a review of the literature in key areas of relevance to our study.

\subsection{Action Recognition}
The field of action recognition as a pivotal task in video understanding, has witnessed significant advancements in recent years, driven by the rise of deep learning techniques and the availability of large-scale video datasets. While early approaches heavily relied on handcrafted features and traditional machine learning algorithms (e.g. \cite{kwapisz2011activity}\cite{ordonez2014home}\cite{poppe2010survey} and \cite{aggarwal2015human}), the emergence of Convolutional Neural Networks (CNNs) revolutionized the field.

CNN-based architectures, such as Two-Stream Networks \cite{liu2021improved}\cite{xiong2020transferable} and 3D Convolutional Networks \cite{yang2019asymmetric}\cite{arunnehru2018human}, have demonstrated remarkable performance in recognizing various human actions, including manipulation actions \cite{cheron2015p}. These models leverage spatiotemporal information in videos to capture the dynamics and visual patterns associated with different actions. Notably, recent trends have focused on incorporating attention mechanisms to attend to salient spatiotemporal regions, enabling more accurate recognition of fine-grained actions \cite{ma2019attnsense}\cite{yin2022novel}.

Moreover, the recognition of manipulation actions has gained specific attention due to its relevance in robotics and human-robot interaction \cite{billard2019trends}\cite{thomaz2016computational}. Researchers have explored specialized architectures that explicitly model hand-object interactions, using both, appearance and motion, cues. These approaches often combine deep learning techniques with hand-crafted features, such as hand pose estimation or object tracking, to improve the recognition accuracy of manipulation actions \cite{oberweger2015hands}\cite{chatzis2020comprehensive}.

However, a notable gap in the literature is the limited focus on bimanual actions, which involve coordinated movements and interactions between both hands. Bimanual actions are prevalent in various domains, including manufacturing, cooking, and assembly tasks \cite{franz2003bimanual}. Understanding and recognizing bimanual actions require capturing the complex dependencies and coordination between the hands, which present unique challenges compared to unimanual actions.

While there have been studies addressing bimanual actions in robotics (e.g. \cite{rakita2019shared}\cite{xie2020deep} and \cite{ureche2018constraints}), the application of deep learning-based action recognition techniques to bimanual actions is relatively unexplored. This provides an exciting opportunity for future research to develop specialized architectures and datasets that specifically target bimanual action recognition. Such advancements would greatly benefit applications in robotics, where understanding bimanual interactions is crucial for human-robot collaboration and task execution \cite{hu2023towards}\cite{galofaro2023bimanual}.

\subsection{Scene Graphs and Graph-based Modeling}

In the realm of video understanding, scene graphs and graph-based modeling have emerged as powerful techniques for capturing the intricate relationships between objects and actions (e.g. \cite{aksoy2010categorizing}\cite{wu2019learning} and \cite{ji2020action}). 

One leading trend in scene graph generation is the integration of deep learning methodologies. Traditional approaches heavily relied on handcrafted features and rule-based heuristics, which limited their scalability and adaptability to diverse video datasets. With the advent of deep learning, researchers have used convolutional neural networks (CNNs) and recurrent neural networks (RNNs) to automatically extract rich visual features and learn complex representations from raw video data \cite{xu2017scene}\cite{yang2018graph}. These deep learning-based techniques have demonstrated remarkable performance improvements in scene graph generation by effectively capturing the semantic relationships between objects \cite{xu2020survey}\cite{teng2022structured}.

A growing trend in scene graph generation involves incorporating contextual information to enhance the quality and relevance of generated graphs. Context plays a pivotal role in understanding visual scenes, as the relationships between objects are influenced by their surrounding environment. Recent methods have explored the integration of contextual cues, such as global scene context \cite{wang2019exploring}, spatial relationships \cite{cong2021spatial}, and temporal dependencies \cite{feng2023exploiting}, to improve the accuracy and contextual understanding of scene graphs. Attention mechanisms and graph propagation techniques have been employed to selectively attend to relevant regions or objects in the scene and propagate information across the graph, capturing higher-order relationships and contextual dependencies \cite{airin2022attention}\cite{nguyen2021oscar}.

Graph-based modeling techniques have also witnessed significant advancements in various aspects of video understanding. Notably, the adoption of graph neural networks (GNNs) has revolutionized the field by extending convolutional operations to graph structures. GNNs enable information propagation and aggregation across nodes in the graph, allowing for the modeling of complex relationships and capturing both spatial and temporal dependencies. By leveraging GNNs in scene graphs, researchers have achieved state-of-the-art results in critical tasks such as action recognition, video captioning, and object detection \cite{khademi2020deep}\cite{li2021bipartite} and \cite{ravichandran2022hierarchical}. The application of GNNs facilitates end-to-end learning and enables comprehensive modeling of visual scenes, contributing to improved video understanding capabilities.

Another noteworthy trend in graph-based modeling is the integration of multi-modal information. Combining visual data with complementary modalities such as textual descriptions or sensor data enhances the understanding and interpretation of video content. For instance, textual information can provide valuable semantic cues that complement the visual scene graph, leading to improved performance in tasks such as video captioning and activity recognition \cite{xu2019scene}\cite{ye2021linguistic}. By using the synergistic fusion of multiple modalities, researchers aim to achieve a more holistic and comprehensive understanding of video content \cite{chen2019synergistic}.

In conclusion, the current trends and methods in scene graph generation and graph-based modeling for video understanding underscore the integration of deep learning methodologies, the incorporation of contextual information, the utilization of graph neural networks, the mitigation of scalability challenges, and the integration of multi-modal data. These advancements have propelled the field forward, enabling more accurate and contextually aware video understanding systems. However, there remain challenges, including handling large-scale scene graphs, improving interpretability, and advancing graph reasoning capabilities, which warrant further research and exploration to unlock the full potential of graph-based modeling in video understanding.

\subsection{Video Description Generation}
Video description generation is a task within the domain of video understanding that aims to automatically generate textual descriptions or captions for video content. This subsection provides a concise overview of the historical developments and the current state of the art in video description generation.
\subsubsection{Historical Developments}
The task of generating textual descriptions for videos has evolved significantly over the years. Early approaches focused on rule-based systems, where predefined templates and handcrafted rules were used to generate descriptions based on visual cues and temporal analysis of the video content (e.g. \cite{kojima2002natural}\cite{nishida1982japanese}\cite{nishida1988feedback} and \cite{lee2008save}). These methods, although limited in their descriptive capabilities, laid the foundation for subsequent research in the field.

With the advent of deep learning and the availability of large-scale video datasets, the field witnessed a shift towards data-driven approaches. Researchers started to employ convolutional neural networks (CNNs) for visual feature extraction and recurrent neural networks (RNNs) for generating sequential descriptions. This approach enabled the modeling of temporal dependencies and improved the descriptive quality of the generated captions.

\subsubsection{Current State of the Art}
In recent years, significant advancements have been made in video description generation, driven by the combination of deep learning techniques \cite{abbas2018video}, multimodal fusion \cite{hori2017attention}, and attention mechanisms \cite{wu2018hierarchical}. State-of-the-art models employ deep neural networks to extract visual features from video frames and textual features from accompanying transcripts or subtitles. These features are then fused and processed through sophisticated architectures to generate coherent and contextually relevant descriptions (e.g. \cite{chen2019deep}\cite{islam2021exploring} and \cite{xu2017learning}).

One notable approach is the Transformer model, originally proposed for natural language processing tasks. The Transformer's self-attention mechanism allows it to capture long-range dependencies and effectively model the relationships between video frames and their corresponding textual descriptions. This has led to remarkable improvements in generating accurate and contextually rich captions for videos (\cite{zhou2018end}\cite{chen2018tvt}\cite{lei2020mart} and \cite{lin2022swinbert}).

Another area of active research is the incorporation of external knowledge sources, such as pre-trained language models and commonsense reasoning, to enhance the descriptive capabilities of video captioning systems. These knowledge-driven approaches aim to generate captions that are not only faithful to the visual content but also demonstrate a deeper understanding of the context and semantics \cite{seo2022end}\cite{yang2023vid2seq}.

Furthermore, researchers are exploring techniques to address challenges in video description generation, such as handling complex scenes, temporal dynamics, and generating captions for videos without accompanying textual annotations. This involves developing novel architectures, relying on unsupervised learning, and integrating multimodal cues, including audio and motion, to provide more comprehensive and informative descriptions \cite{iashin2020better}\cite{iashin2020multi}\cite{nagrani2022learning}.

\subsection{Gap Analysis and Our Proposed Work}
In this section, we would like to point out key gaps in the state of the art that our research aims to address.

A significant gap within action recognition literature lies in the limited exploration of bimanual actions, which involve detailed interactions between both hands. Despite the complexity and common occurrence of such actions in various sectors, they have been largely disregarded in deep learning-based action recognition. Our research proposes an innovative methodology for recognizing and understanding bimanual actions.

In terms of Scene Graphs and Graph-based Modeling, the handling of large-scale scene graphs and the furthering of graph based reasoning capabilities are identified as challenges that demand additional research. Despite the progress in the field, scalability and interpretability of scene graphs still remain as hurdles. Our research contributes to this area, introducing methods to manage large-scale scene graphs effectively, and providing enhanced reasoning techniques to improve the interpretability of these graphs.

For video description generation, a comprehensive approach to describe complex video content is another hurdle not yet  surmounted by the existing body of work. Challenges lie in handling complex scenes, temporal dynamics, and in generating descriptions for videos that lack textual annotations. Our research seeks to make a contribution in this area, adopting a comprehensive approach that utilises contextual cues and innovative architectural designs to generate relevant descriptions for complex video content.



\begin{figure*}[!h]
    \centering
    \includegraphics[width=0.99\textwidth]{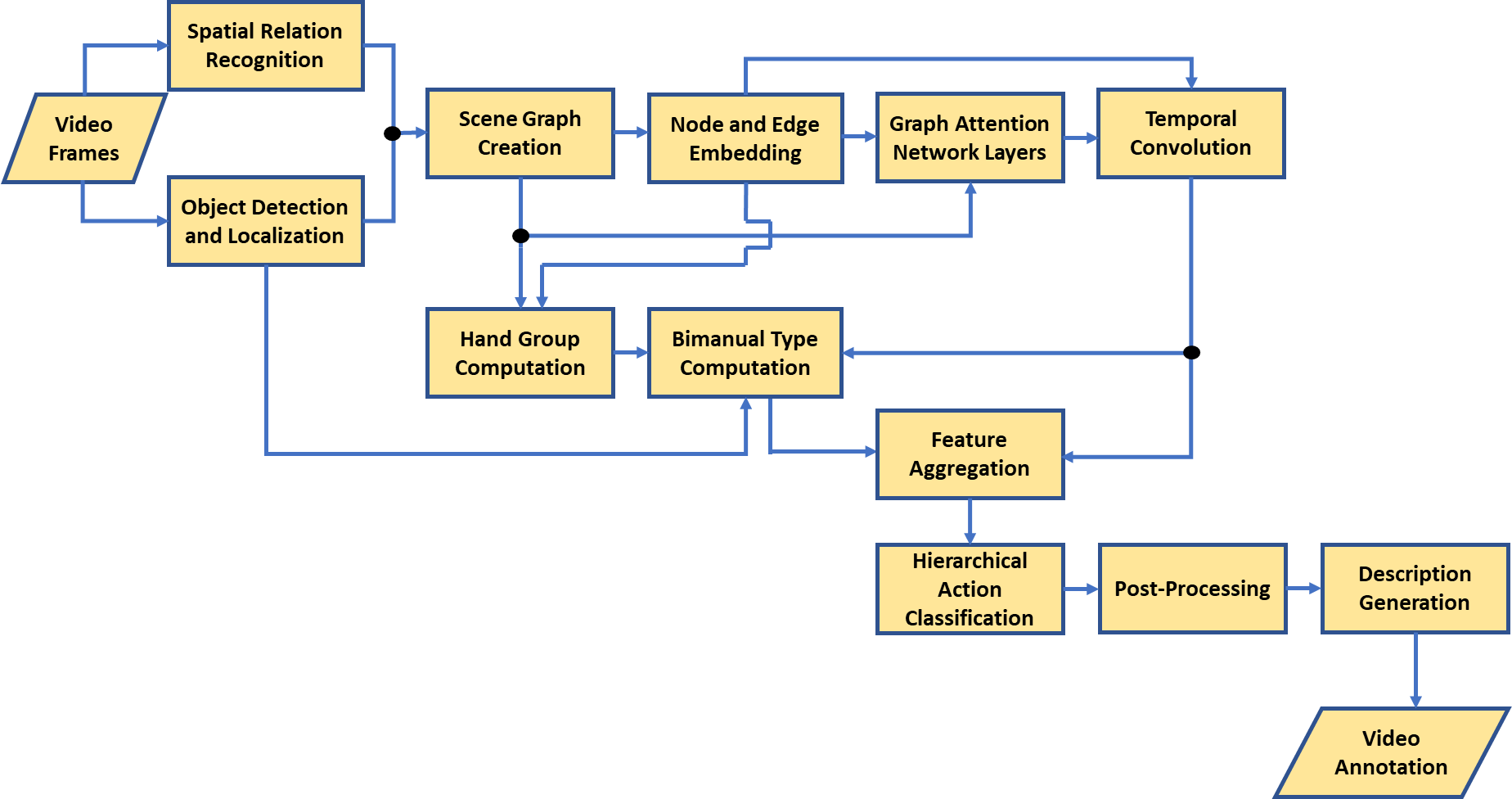}
    \caption{The flow diagram of BiGNN framework }
    \label{Fig.overview}
\end{figure*}

\section{METHODS}
\label{methods}

In this section, we present the Bimanual Graph Neural Network (BiGNN), a framework for recognizing bimanual actions and generating natural language descriptions of these actions at multiple levels of detail using graph neural networks. BiGNN comprises several interconnected components, each serving a unique role in the overall action recognition and description process. Note the most of the finer algorithmic details are found in the Supplementary Material including all learning aspects.

As shown in Figure \ref{Fig.overview}, the first step of BiGNN involves recognizing objects and spatial relations to create a scene graph representation of the bimanual action. This scene graph is then processed through an embedding layer, and the resulting features are utilized by a modified Graph Attention Network (GAT) architecture, designed to extract important features while taking spatial information into account. The extracted features are then passed into a temporal convolutional network (TCN) to identify the temporal dependencies between the video frames. The roles of the different hands are also extracted from the video frames to further enhance the recognition of bimanual actions. Finally, the hierarchical processing of the BiGNN framework comes into play, where the action is recognized at different levels of detail. As will be shown below, this way BiGNN generates human-like sentences describing the action at different levels of detail, from high-level overviews to more detailed descriptions of the individual objects and their interactions, which enables understanding of the bimanual action at different level of description granularity.
In the following the different components will be described.

\subsection{Pre-processing and Scene Graph Creation}
\label{graph}

The different pre-processing steps are crucial components of the Bimanual Graph Neural Network (BiGNN) framework, responsible for detecting and localizing objects in the video frames, and recognizing their spatial relations. 

Pre-processing begins with object detection, utilizing a pre-trained deep learning model to detect and localize objects within each video frame. Following object detection, the spatial relations between objects are recognized and represented as a graph. Each object is represented as a node in the graph, and the edges between nodes represent the spatial relations between objects, such as proximity or relative position. 

In the following sections, we will discuss the details of pre-processing, including object detection, spatial relation recognition, and scene graph creation.

\subsubsection{Object Detection}
\label{object}

The initial stage of our pre-processing pipeline involves the 2D analysis of RGB images to detect objects using the You Only Look Once (YOLO) algorithm \cite{redmon2018yolov3}. YOLO has been trained on the labeled objects within our dataset, enabling it to accurately identify and localize objects of interest. Additionally, we utilize OpenPose \cite{cao2021openpose} to detect human hands, which provides key points representing hand positions. From these key points, we calculate 2D bounding boxes for each hand. This stage produces a list of 2D bounding boxes, encompassing both the objects detected by YOLO and the hands detected by OpenPose. It is worth noting that the hands are treated as any other object in the subsequent stages of our pipeline. Note that BiGNN is not bound to using these modules (YOLO, OpenPose) but they fully suffice for our purpose.

\subsubsection{Spatial Relation Recognition}
\label{SR}

The second stage of this step involves 3D pre-processing of the data, which is specifically performed on 3D datasets (for 2D datasets see end of this subsection). During this stage, the information obtained from the first step is used alongside the point clouds generated from the depth images of the 3D dataset to acquire precise 3D bounding boxes for the objects. If the pre-segmented information of the objects is already available in the dataset, we generate their corresponding 3D models by creating the 3D surrounded convex hulls. In cases where the dataset lacks pre-segmented objects, their corresponding Axis Aligned Bounding Box (AABB) is utilized to model the objects in 3D space. When depth images are unavailable in the dataset, the approach proposed in \cite{birmingham2018adding} is employed to add a third dimension to the 2D images, allowing for the generation of 3D bounding boxes. Finally, the spatial relations between each pair of objects are computed using the well-established methods from \cite{aksoy2011learning}\cite{ziaeetabar2017semantic}\cite{ziaeetabar2018recognition}. The methodology of computing these spatial relations has been explained in \cite{ziaeetabar2018recognition}.


The list of our defined spatial relations SR includes static spatial relations ($SSR$) pertaining to object positions, comprising the thirteen items. In the original framework \cite{ziaeetabar2017semantic} also dynamic spatial relations pertaining to object movement ($DSR$) had been defined but are not used here. SSRs are given as:

\begin{equation}
\label{relations}
\begin{multlined}
\text{SSR} = \{\text{contact, above, below, around, left, right, front, behind,}\\ 
\text{contain, partial contain, within, partial within, cross\}}\\ 
\end{multlined}
\end{equation}

\subsubsection{Spatio-Temporal Scene Graph Construction}
\label{representation}

In our approach, each frame of the video contains a scene graph, which is represented as an undirected graph $\mathcal{G}=(\mathcal{V},[\mathcal{E}_1,\mathcal{E}_2,\mathcal{E}_3])$. The set $\mathcal{V}$ consists of node attributes $\tilde{h}$, each of which represents an object in the scene, $\mathcal{E}$ are edges between those nodes (see below).

Node attributes $\tilde{h}$ for node $i$ are given as a mixed tuple $\tilde{h}_i=\{n_i, \eta(p_i), \eta(s_i)\}$, where $n$ is a string (label of the object), and $p$ as well as $s$ are position and speed, encoded by numerical values. The operator $\eta$ refers to a normalization function used to keep values bounded. We used, as also commonly done: $\eta(y) = (y - \text{mean}(y)) / \text{std}(y)$.
Below we will describe how to transform the mixed tuple into a fully numerical feature vector.

The set $[\mathcal{E}_1,\mathcal{E}_2,\mathcal{E}_3]$ consists of undirected edges, where $\mathcal{E}_1$ represents \textit{spatial relation} attributes $\tilde{e}$ between two objects in one given video frame, $\mathcal{E}_2$ refers to \textit{temporal-relation} attributes $\tilde{v}$ between objects from the last ($t-1$) to the current ($t$) movie frame, and $\mathcal{E}_3$ represents \textit{actions} $\tilde{a}$ that can be performed given an object-pair in the scene. Note that temporal relations will eventually be defined by object-object distance changes and only be used much later (see Section on TCNs, below). Until then we will only deal with spatial- and action-attributes.

Hence, for the spatial attributes $\tilde{e}$ of an edge of type $\mathcal{E}_1$ between nodes $i$ and $j$, we define accordingly the tuple: $\tilde{e}_{i,j}=\{y_{i,j}, \eta(p_i), \eta(p_j)\}$. Here the string variable $y$ refers to the spatial relation type (see Eq.~\ref{relations}) encoded by the edge, and, as before, $p$ refers to the position of an object (=node). String entries $y$ are computed for the spatial edges from the static spatial relations (SSR, see Eq.~\ref{relations} and \cite{ziaeetabar2017semantic}).

The action-edge attributes $\tilde{a}$ of an edge of type $\mathcal{E}_3$ between nodes $i$ and $j$ are defined for each possible action $k$. Hence, each action index $k$ corresponds to a specific action category in our dataset(s). These edges, thus, represent the actions that can be performed between pairs of objects. For example, if one object is ``knife" and the other is ``fruit," the object-action edge could correspond to the action of ``cutting."  Each action attribute $\tilde{a}_{i,j,k}$ is, thus, represented by a tuple $\{s_{i,j,k}, \eta(p_i), \eta(p_j)\}$, where: 
$s_{i,j,k}$ represents the specific action's importance score between nodes $i$ and $j$ for the given action index $k$. This importance score quantifies the relevance of the action between the two nodes and is learned using the GAT (see below). It captures how significant a particular action is in the context of the interaction between nodes $i$ and $j$. The variables $\eta(p_i)$ and $\eta(p_j)$ represent, as before, the spatial positions of nodes $i$ and $j$, respectively. These attributes combine the action-related information (importance scores), with the spatial context of the nodes, enabling the graph attention mechanism to capture both action-specific relationships and spatial dependencies during the graph attention computation. 
The model can now handle multiple possible actions between objects, capturing the diversity and complexity of bimanual actions in the scene graph representation.

\begin{figure}[]
    \centering    \includegraphics[width=0.8\textwidth, height=0.8\textheight]{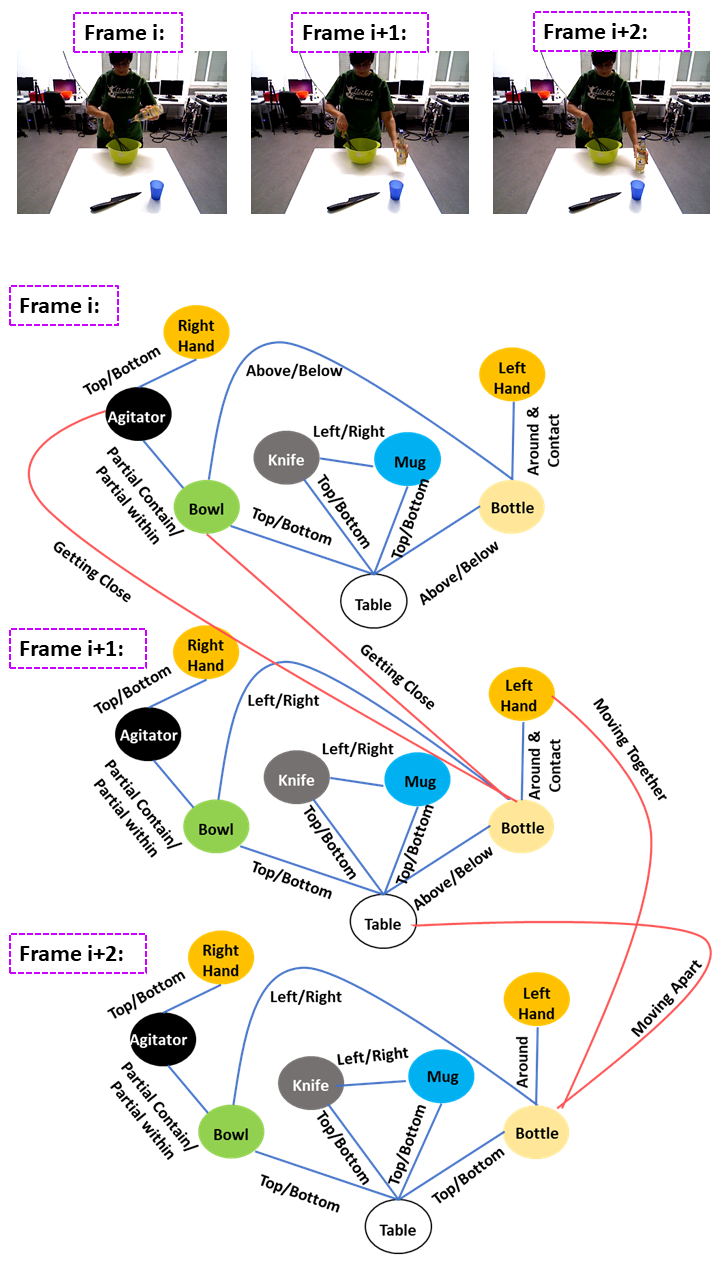}
    \caption{An example of a scene in the KIT bimanual action dataset \cite{dreher2020learning} and its related scene graphs at three consecutive frames. In this figure blue and red lines indicate spatial and temporal edges respectively. }
    \label{fig:scene_graph}
\end{figure}

\begin{figure}[]
    \centering
   \includegraphics[width=0.99\textwidth]{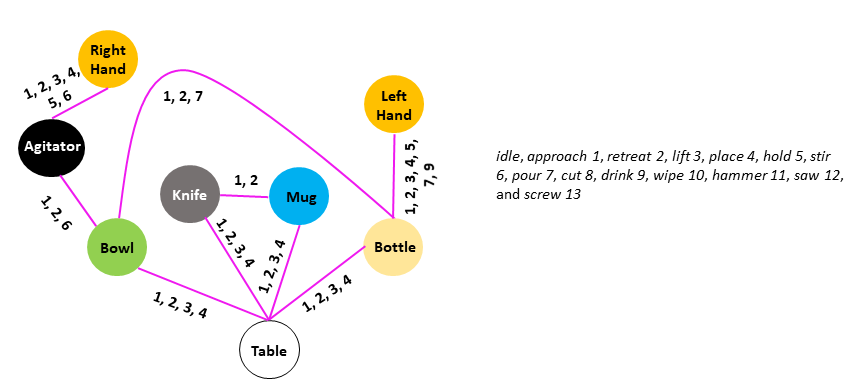}
    \caption{Object-action edges related to the scene graph in Figure~\ref{fig:scene_graph}. The purple lines show object-action edges. In the depicted dataset, the available actions are labeled as: approach 1, retreat 2, lift 3, place 4, hold 5, stir 6, pour 7, cut 8, drink 9, wipe 10, hammer 11, saw 12, and screw 13. The numbers written above the purple edges represent the possible actions that can occur between those two node objects.}
    \label{fig:actions}
\end{figure}

A sample of three consecutive video frames in a bimanual action manipulation and their corresponding scene graphs is shown in Figures \ref{fig:scene_graph} and \ref{fig:actions}. In this figure blue edges represent spatial relations (e.g. above, left...) between each pair of objects in a frame that their distance is less than a threshold and red edges demonstrate temporal relations between those through continuous frames. For example, in the second frame, the left hand lifts the bottle from the table and holds it above the bowl which means the bottle is moving apart from the table (their corresponding SSR changes from Top/Bottom to Above/Below) and is getting close to the bowl (their corresponding SSR changes from Left/Right to Above/Below). To avoid cluttering the figure, only a subset of edges are shown.\\

\subsection{Node and Edge Feature Encoding}
\label{embedding}
After the creation of the scene graph in Section \ref{graph}, we proceed to transform the node as well as spatial- and action-edge attributes into a numerical encoding. For this, we utilize the Word2Vec algorithm \cite{mikolov2013efficient}. In doing so we obtain now fully numerical feature vectors $\vec{h}_i$ for nodes and $\vec{e}_{i,j}$ as well as $\vec{a}_{i,j,k}$ for the edges, resulting in  $\mathbf{\hat{H},~\mathbf{\hat{E}}}$ and $\mathbf{\hat{A}}$. Note that, for simplicity, we will use the word ``matrix" for all such entities even if their dimension is larger than 2 (where ``tensor" should be used instead).

The preliminary feature matrices are then augmented using linear layers represented by weight matrices $\mathbf{L}_H$, $\mathbf{L}_E$, and $\mathbf{L}_A$, respectively. Please see Supplementary Material for the procedures for computing $\mathbf{L}$.
Such linear layers allow capturing more complex patterns and relationships in the data. They operate on the preliminary feature matrices $\mathbf{\hat{H}}$, $\mathbf{\hat{E}}$, and $\mathbf{\hat{A}}$, transforming them into the final feature matrices $\mathbf{H}$, $\mathbf{E}$, and $\mathbf{A}$, respectively. The linear transformation is mathematically defined as follows:
\begin{eqnarray}
\mathbf{H} &=& \mathbf{L}_H \otimes \mathbf{\hat{H}} \\
\mathbf{E} &=& \mathbf{L}_E \otimes \mathbf{\hat{E}}  \label{encoding1}\\
\mathbf{A} &=& \mathbf{L}_A \otimes \mathbf{\hat{A}}
\label{encoding2}
\end{eqnarray}
In this process, $\otimes$ represents the Hadamard product. These final feature matrices $\mathbf{H}$, $\mathbf{E}$, and $\mathbf{A}$ now contain encoded information about the nodes, spatial and temporal edges, and object-action edges, which are then used in subsequent stages of our method for action description.

\begin{figure}[]
    \centering
    \includegraphics[width=0.98\textwidth]{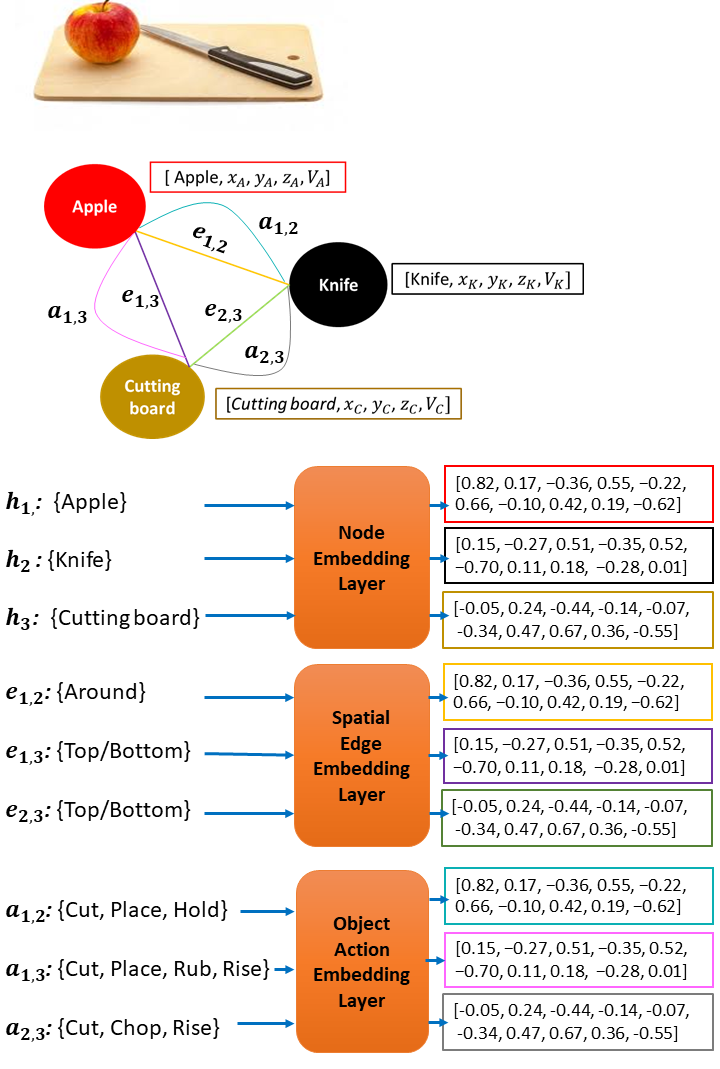}
    \caption{Illustration of the node and edge feature encoding process in a scene graph, where object attributes and spatial relations are represented as continuous vectors using an embedding layer. $x$, $y$, $z$ are the Cartesian coordinates of an object, and $v$ represents its velocity.}
    \label{Fig.embedding}
\end{figure}

Figure \ref{Fig.embedding} illustrates an example scene with three objects: an apple, a knife, and a cutting board. The apple and the knife are placed on the cutting board, and the knife is positioned around the apple. The corresponding scene graph and the embedded node and edge feature matrices.\\

\subsection{Graph Attention Network (GAT): general definition}
\label{GAT}

We use GATs in two sub-processes, below. Hence, we will define a GAT first in general terms and eventually augment this definition in the respective sub-sections as needed.

Let $\vec{h}_{i}^{(l)}$ be a feature vector of node $i$ at the $l$-th layer of a GAT. The GAT model computes new features for each node by aggregating features from its neighbors weighted by attention scores $\alpha$:
\begin{eqnarray}
\vec{z}_{i}^{(l)} &=&\mathbf{W}^{(l)} \vec{h}_{i}^{(l)}, \label{defZ} \label{def-z}\\
f_{ij}^{(l)} &=& \text{LeakyReLU}\left(\vec{d}^{(l)}\left[ \vec{z}_{i}^{(l)}, \vec{z}_{j}^{(l)} \right]^T\right), \label{score}\\
\alpha_{ij}^{(l)} &=& \frac{\exp(f_{ij}^{(l)})}{\sum_{k\in N_{i}}\exp(f_{ik}^{(l)})}, \\
\vec{h}_{i}^{(l+1)} &=& \sigma\left(\sum_{j\in N_{i}} \alpha_{ij}^{(l)} \vec{z}_{j}^{(l)}\right), \label{nextstep}
\end{eqnarray}
where $\mathbf{W}^{(l)}$ and $\vec{d}^{(l)}$ are learnable (see Supplementary Materials), $\text{LeakyReLU}$ is the leaky rectified linear unit activation function, $\sigma$ is the activation function (e.g., sigmoid or softmax), and $N_{i}$ is the set of neighbors of node $i$. The notation $[\dots,\dots]$ means concatenation. 

Each attention score $\alpha_{i,j}^{(l)}$ measures the importance of node $j$ to node $i$ at layer $l$. 

In total, we have implemented three layers. Hence, our GAT-process ends at $\vec{h}^{(3)}$. We call this feature vector ``final" and abbreviate it with $\vec{h}^f$.

Note that in section \ref{GAT_A} we augment this general GAT process by replacing node features $h$ with action featuers $a$ in Eq.~\ref{def-z}. In section \ref{GAT_H} we modify Eq.~\ref{score} in several ways defining three more GAT processes.


\subsubsection{Preprocessing action features $a$ using GAT}
\label{GAT_A}
Action features were defined above as $\vec{a}_{i,j,k}$, but --- for simplicity --- we are dropping index $k$ at the action-edges and emphasize that all further processing is performed for each action $k$ in the same way.

We use a first GAT process to perform aggregation of  $\vec{a}^{(l)}_{i,j}$ across the GAT layers.  By aggregating information from each layer, the final feature vector $\vec{a}_{i,j}$ encodes more reliably how the edge between nodes $i$ and $j$ represents a specific action performed by those objects.\\

\noindent
\textbf{Initialization of $\vec{a}_{i,j}^{(0)}$:}\\
As a common approach, the initial feature vector $\vec{a}_{i,j}^{(0)}$ is randomly initialized using a Gaussian distribution with mean 0 and a small standard deviation. This initialization ensures that the initial features are centered around zero and have small magnitudes, which aids in the initial stages of learning.
The initialization process can be expressed as: $\vec{a}_{i,j}^{(0)} \sim \mathcal{N}(0, \sigma^2)$, where $\sigma$ is a small positive value representing the standard deviation of the Gaussian distribution.

By using this approach, the initial edge features are set to random values, which allows the model to adapt and update these features based on the patterns and information present in the data during training. Although the exact values of the initial edge features are not critical, this initialization provides a reasonable starting point for the learning process and for this we replace $\vec{h}_{i}^{(0)}$ (Eq.~\ref{defZ}) with $\vec{a}_{i,j}^{(0)}$.\\

\noindent
\textbf{Aggregation across layers:}\\
After the GAT-learning is completed we combine the information from all layers and obtain a single representation for each action edge $\vec{a}_{i,j}$ by an aggregation step. The aggregation of $\vec{a}_{i,j}$ across layers is typically performed using a pooling operation, such as summation or mean pooling. We use summation:
\begin{equation}
\vec{a}_{i,j} = \frac{1}{L} \sum_{l=1}^{L} \vec{a}_{i,j}^{(l)}
\end{equation}
where $L$ is the total number of GAT layers, where we use $L=3$.\\

\subsubsection{Hierarchical GAT Levels}
\label{GAT_H}

The above definitions and the preprocessing of the actions allows now to introduce a key feature of our approach. Different from conventional GATs we are implementing a new aspect by creating a 3-level hierarchical GAT process, which allows us to recognize actions at multiple levels of detail. Specifically, we define three levels of GAT: 1) \textbf{object-level}, 2) \textbf{single hand action-level}, and 3) \textbf{bimanual action-level} annotated with superscripts $.^{(1)},.^{(2)}, .^{(3)}$ in the following.

At the object-level, our GAT learns relationships between individual objects in the scene graph, such as which objects are near each other or which objects are frequently used together. For example, the GAT might learn that the knife is frequently used to cut fruit, or that the bowl is frequently used to hold chopped fruit.

At the single hand action-level, we use the output of the object-level GAT to identify which objects are being manipulated by each hand. We can then use this information to recognize the actions of each individual hand. For example, the GAT might recognize that the left hand is holding a knife and the right hand is holding a piece of fruit, and that the action being performed is ``slice".

At the bimanual action-level, we use the output of the action-level GAT to identify which actions involve multiple hands working together. We can then use the output of the object-level GAT to identify which objects are being manipulated by each hand in the bimanual action. For example, the GAT might recognize that both hands are being used to mix the chopped fruit together.

To formalize this we create three levels, where a higher level uses the output of the previous level to identify and analyze more complex relationships between nodes in the scene graph. Remember that, in the following, superscripts refer to GAT levels and not (as above in Eqs.~\ref{defZ}-\ref{nextstep}) to the layers of the GAT.

We define $\mathbf{H}^{(0)}$ and $\mathbf{E}^{(0)}$ as embedded original node and edge feature matrices, then: 
\begin{equation}
\mathbf{G}^{(1)} = \text{GAT}(\mathbf{H}^{(0)}, \mathbf{E}^{(0)})
\label{G1}
\end{equation}
This equation represents the first GAT level (object level), which is applied to the initial node feature matrix $\mathbf{H}^{(0)}$ and the spatial edge feature matrix $\mathbf{E}^{(0)}$.
To get this we expand the right side of Eq.~\ref{score} with spatial edge and action-edge features such that it reads now:
\begin{equation}
f^{(1)}_{i,j} = \text{LeakyReLU}\left(\vec{d}\left[ \vec{z}_{i}, \vec{z}_{j},\mathbf{W}_e \vec{e}_{i,j}, \mathbf{W}_a \vec{a}_{i,j} \right]^T\right),
\label{modscore2}
\end{equation}
where indices $e$ and $a$ depict that the (learnable) matrices $\mathbf{W}$ correspond to edges $e$ and $a$. The notation $[ \dots,\dots ]$ represents concatenation.

In this step, the GAT uses the original node and edge features to compute the attention scores $f_{i,j}$ (as described in Eq.~\ref{score}) between each pair of nodes $i$ and $j$. The output of this step is the updated  node feature matrix $\mathbf{G}^{(1)}$, which is obtained according to the node and edge information at Level 1 (object level).

As a result all feature matrices above level 0 now contain mixed node and edge information. Next we define:
\begin{equation}
\mathbf{G}^{(2)} = \text{GAT}(\mathbf{G}^{(1)}, \mathbf{B}^{(2)}, \mathbf{F}^{(1)})
\label{G2}
\end{equation}
We explain the new terms $\mathbf{G}^{(1)},\textbf{B}^{(2)}$, and $\mathbf{F}^{(1)}$ in order of appearance. They are used to define $f_{ij}^{(2)}$ (compare to Eq.~\ref{score}) as:
\begin{equation}
f_{ij}^{(2)} = \text{LeakyReLU}\left(\vec{d}\left[ \mathbf{G}^{(1)}\vec{h}_{i}^{(0)}, \mathbf{G}^{(1)}\vec{h}_{j}^{(0)}, \mathbf{W}_e\vec{e}_{i,j}, \mathbf{W}_a\tilde\vec{a}_{i,j},  \mathbf{F}^{(1)}\vec{\gamma} \right]^T\right)
\label{deff2-correct}
\end{equation}
\noindent
\underline{On $\mathbf{G}^{(1)}$:} In the equation for $f_{ij}^{(2)}$, $\mathbf{G}^{(1)}$ refines the initial node embeddings. It transforms the initial features using learned weights and captures the collaborative relationships between nodes at the first GAT layer. Therefore, the refined embeddings, $\mathbf{G}^{(1)}\vec{h}_{i}^{(0)}$ and $\mathbf{G}^{(1)}\vec{h}_{j}^{(0)}$, encode complex node interactions based on both intrinsic node properties and their relationships with neighboring nodes.\\

\noindent
\underline{On $\mathbf{B}^{(2)}$:} The attention matrix $\mathbf{B}^{(2)}$ is used to weigh $\vec{a}_{i,j}$ to obtain $\tilde{\vec{a}}_{i,j}$ in Eq.~\ref{deff2-correct}. For this we define the elements of the matrix $\mathbf{B}^{(2)}$ as $\beta_{i,j}$ given by the following softmax operation:


\begin{equation}
\beta_{i,j} = \frac{\exp((\mathbf{W}^{(2)} \cdot \mathbf{G}^{(1)}_j)  \cdot ( \mathbf{U}^{(2)}\cdot \mathbf{G}^{(1)}_i)^T)}{\sum_{k \in \mathcal{N}(i)} \exp((\mathbf{W}^{(2)} \cdot \mathbf{G}^{(1)}_k)  \cdot ( \mathbf{U}^{(2)}\cdot \mathbf{G}^{(1)}_i)^T)}
\label{softmax1}
\end{equation}

In Equation \ref{softmax1}, $\mathbf{U}^{(2)}$ is a learnable weight matrix specific to the second GAT layer. It is used to transform the node features of neighboring nodes \( \mathbf{G}^{(1)}_j \) before calculating the attention scores. $\mathbf{W}^{(2)}$ is a transformation matrix for node features at Layer 2, refining the representations based on the outputs of Layer 1. See Supplementary Material for details.



In this equation, $\mathcal{N}(i)$ denotes the set of neighbors of node $i$. The outcome, $\beta_{i,j}$, provides another normalized attention score that captures the significance of the information from node $j$ for node $i$ but here concerning action edges which is computed by element-wise multiplication given as:
\begin{equation}
\tilde{\vec{a}}_{i,j} = \beta_{i,j}  \vec{a}_{i,j}
\label{a-mod}
\end{equation}
$\mathbf{B}^{(2)}$ is called attention matrix at the single-hand action-level. It determines the importance of each action edge for recognizing single-hand actions. \\

\noindent
\underline{On $\mathbf{F}^{(1)}$:} The feature matrix $\mathbf{F}^{(1)}$ originates from the object-level-1 GAT layer, which operates on the embedded node feature matrix $\mathbf{H}^{(0)}$. It is column-wise defined as:
\begin{equation}
\label{defE1}
\mathbf{F}_c^{(1)} = \left[ \alpha_{ij}^{(1)} \vec{h}_{i}^{(0)} \right]_c
\end{equation}
where index $c$ stands for column $c$ of the matrix. Thus, $\mathbf{F}^{(1)}$ captures the spatial relationships between objects in the scene graph, and it is obtained by weighting the embedded node feature matrix $\mathbf{H}^{(0)}$ with attention weights $\alpha$ from $\mathbf{G}^{(1)}$ calculated based on the relationships between nodes at the object-level GAT layer.

To correctly define Eq.~\ref{deff2-correct}, $\mathbf{F}^{(1)}$ is multiplied with $\vec{\gamma}$. The vector $\vec{\gamma}$ is a learnable parameter and serves as a weighting mechanism that allows us to control the influence of the matrix $F^{(1)}$ on the attention score calculation. It determines the relative importance/contribution of each column of $F^{(1)}$ to the attention calculation. This introduces a fine-grained level of control over how the matrix $F^{(1)}$ impacts the attention mechanism.

Accordingly the next layer is defined by:
\begin{equation}
\mathbf{G}^{(3)} = GAT(\mathbf{G}^{(2)}, \mathbf{B}^{(3)}, \mathbf{F}^{(2)}),
\end{equation}
with the same definitions as above. GAT layer 3 then captures bi-manual manipulations.

Hence, the 3-level GAT process produces as final outputs $h^{(1),f},h^{(2),f},h^{(3),f}$.

This hierarchical GAT framework enables us, thus, to recognize actions at different levels of detail. By recognizing relationships between objects, individual hands, and both hands, we can, thus, describe actions with greater specificity and accuracy.

\begin{figure}[!h]
\centering
\includegraphics[width=0.98\textwidth]{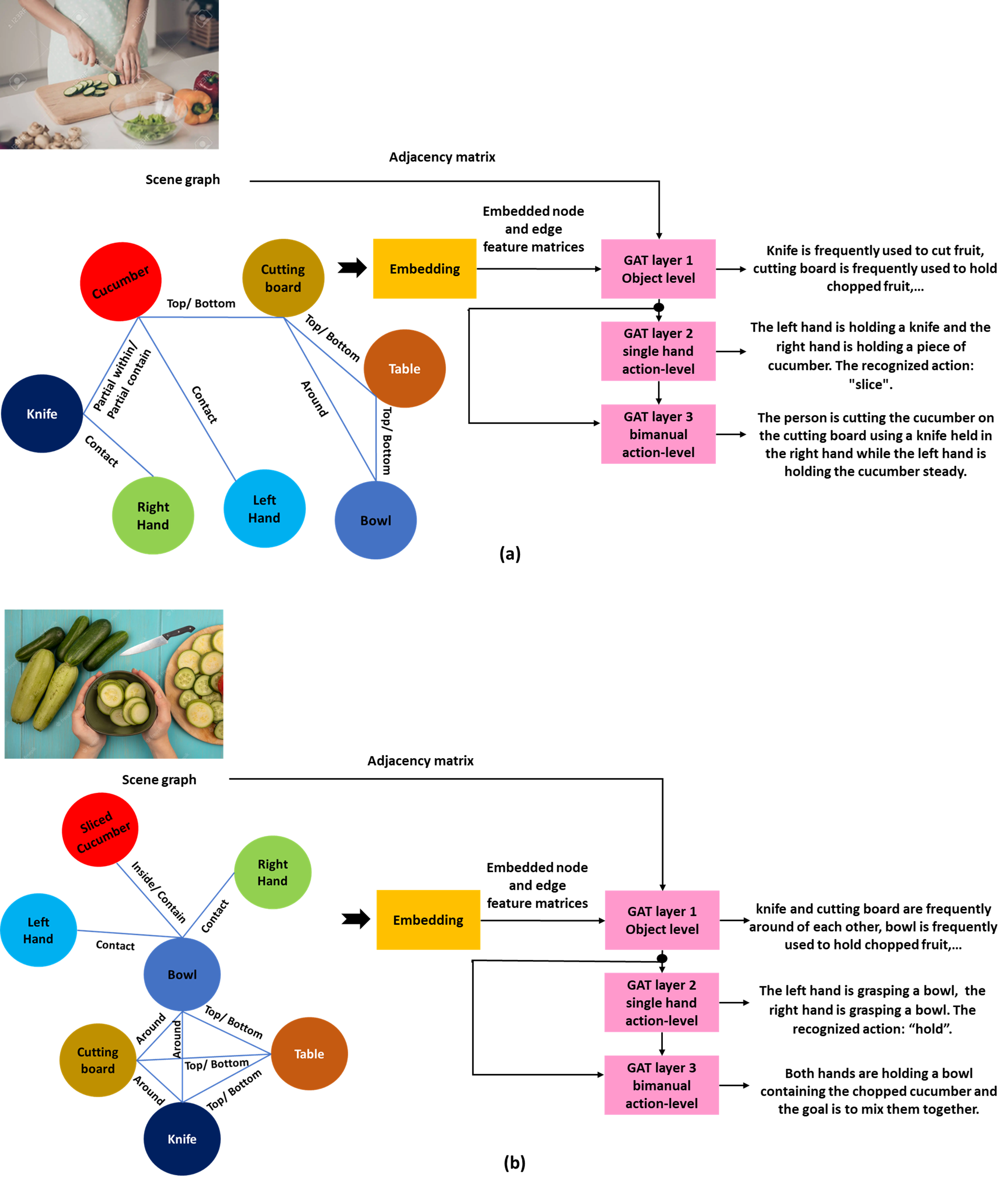}
\caption{Illustration of the use of GAT layers with edge features on two scenes of a scenario (a) chopping a cucumber and (b) mixing the chopped pieces of cucumber in a bowl}
 \label{Fig.gat}
\end{figure}

Figure \ref{Fig.gat} depicts two instances of this scene along with their corresponding scene graphs and the respective output of each GAT layer involved.

\subsection{TCN-Based Spatio-Temporal Processing}
\label{TCN}

In this section, we extend the framework established in Section \ref{GAT}, by introducing Temporal Convolutional Networks (TCNs) along with a sliding window procedure to allow addressing ``time", too. This integration empowers our model to capture both short-term and long-term dependencies in the temporal evolution of embedded feature vectors, enabling spatio-temporal reasoning.

To achieve this, we had above introduced three levels of GAT layers: object-level, single hand action-level, and bimanual action-level. Each level is now paired with a corresponding TCN layer to enable temporal reasoning. Specifically, the output of the object-level GAT layer, denoted as $\mathbf{G}^{(1)}$, serves as the input to the TCN layer at the object-level. This TCN layer processes the temporal evolution of embedded node features, capturing short-term and long-term dependencies within the individual objects' relationships. Similarly, the output of the single hand action-level GAT layer, denoted as $\mathbf{G}^{(2)}$, is fed into the TCN layer at the single hand action-level, which focuses on recognizing actions performed by each hand independently. Finally, the output of the bimanual action-level GAT layer, denoted as $\mathbf{G}^{(3)}$, is processed by the TCN layer at the bimanual action-level, allowing the model to recognize actions that involve both hands working together. This hierarchical combination of GAT and TCN layers enables our framework to perform comprehensive spatio-temporal reasoning and produce detailed action descriptions.

Temporal Convolutional Networks (TCNs) are designed for processing sequential data. To achieve this, TCNs use convolutional filters that are shared across the input sequence, enabling information to propagate through the network without degradation. This aspect is combined with a sliding window approach where the TCN applies a fixed-duration  kernel, with temporal kernel size denoted by $K$, to the input sequence. At each time step, the convolutional filter slides across the input sequence, capturing information within a local temporal context of size $K$ video frames. For our different data sets, we use different kernel sizes adapted to the characteristics of the respective data (for values of $K$ see Supplementary Material).

To enhance performance, furthermore we are stacking multiple TCN layers on top of each other. This way, the model can capture increasingly complex temporal dependencies. Each TCN layer extracts higher-level features from the output of the previous layer, allowing the network to recognize longer-term patterns in the data.

\subsubsection{Definition of the TCN Inputs}
The TCN receives input $X = \{\vec{x}_1, \vec{x}_2, \dots, \vec{x}_T\}$ defined in the following.\\

\noindent
\textbf{\underline{Pooling spatial information:}} For each node $i$, the enhanced final feature vector $h^f$, resulting from the GAT layers, is combined with the embeddings of its associated spatial and action edges. This can be represented as:
\begin{equation}
\text{Agg}(\vec{h}^f_i) = \max \left(\vec{h}_i^f, \max_{j} (\mathbf{E}_{i,j}), \max_{j} (\mathbf{A}_{i,j}) \right)
\label{def-X}
\end{equation}
We use maxpooling here because it captures dominant features. This is particularly useful in action recognition because it brings the most dominant and ``actionable" information to the forefront. In dynamic scenes, where multiple activities might be taking place, it's the dominant or significant (inter-)actions that often define the overall action. 

After node-level aggregation, the next step is to aggregate across all nodes to obtain a comprehensive representation for the entire frame at time $t$.
\begin{equation}
\vec{x}^*_t = \max_{i \in \{1,2,...,n\}} \text{Agg}(\vec{h}^f_i)
\end{equation}
This hierarchical method using max-pooling ensures that the final representation $\vec{x}^*_t$ for each frame captures the most dominant features from both nodes and edges, but temporal information is still missing and will be added by the following steps.\\


\noindent
\textbf{\underline{Incorporating temporal information:}} Temporal information between two objects $i$ and $j$ over consecutive frames is defined as:
\begin{equation}
\Delta_{i,j}^{t,t+1} = r_{i,j}^{t+1} - r_{i,j}^{t}
\end{equation}
where $r_{i,j}^{t}$ represents the relative position of objects $i$ and $j$ at frame $t$.  For a given frame $t$, the enriched feature vector that aggregates both node information and temporal information is:
\begin{equation}
\vec{x}_t = \vec{x}^*_t \oplus \sum_j w_{i,j} \cdot \Delta_{i,j}^{t,t+1}
\end{equation}
where $\oplus$ denotes element-wise summation and $w_{i,j}$ signifies the influence of the positional change between objects $i$ and $j$, which we define by an inverse distance relation given as: $w_{i,j}=1/(1+\delta_{i,j})$, where $\delta$ is the distance between both objects.

Through this method, the frame-level feature vectors encapsulate the dynamic interplay of object positions and their relationships across the video and this way we get as one input to the TCN: $X = \{\vec{x}_1, \vec{x}_2, \dots, \vec{x}_T\}$. Note that we segment these vectors into smaller clusters by a windowing process where we use a window size of $30$ frames and a windowing overlap of $10$ frames to divide the video into shorter chunks. Thus, each frame is part of multiple windows which improves the method's capability to capture short-term and long-term contextual dependencies. To handle boundaries, we use padding of the temporal sequence with zero vectors to extend the sequence of both sides.

Thus, with this sliding window procedure, we transform the original temporal sequence $X$ into a collection of overlapping input sequences, which are subsequently fed into the TCNs for action recognition.\\

\noindent
\textbf{\underline{Output of the TCN}}
We define the resulting output vectors and their matrix as $\vec{g}^{(k)}$ and $\mathbf{G}^{(k)}$, for further processing as described next, where $k$ refers to the GAT level.\\\\


\begin{figure}[!h]
\centering
\includegraphics[width=0.98\textwidth]{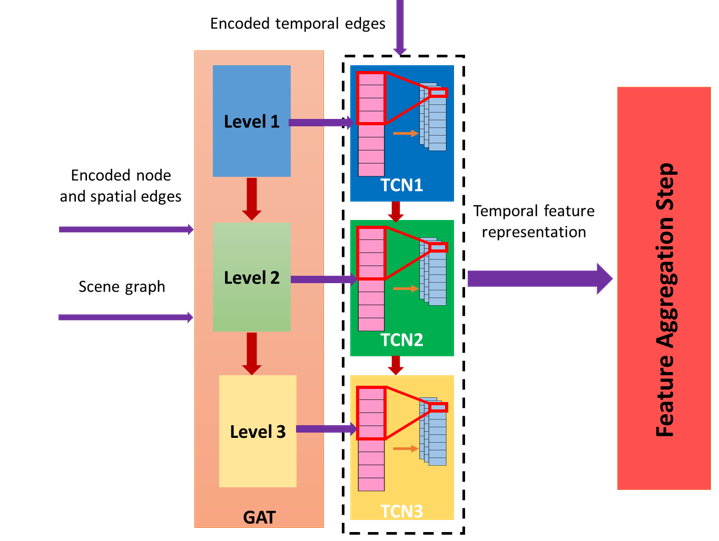}
\caption{A schematic of the temporal convolution step which incorporates dynamic embedded edges of the scene graph into our framework.}
 \label{Fig.TCN}
\end{figure}

\subsection{Rule-based Computation of Hand Groups}
\label{hand_group}
\begin{figure}[!h]
\centering
\includegraphics[width=0.98\textwidth]{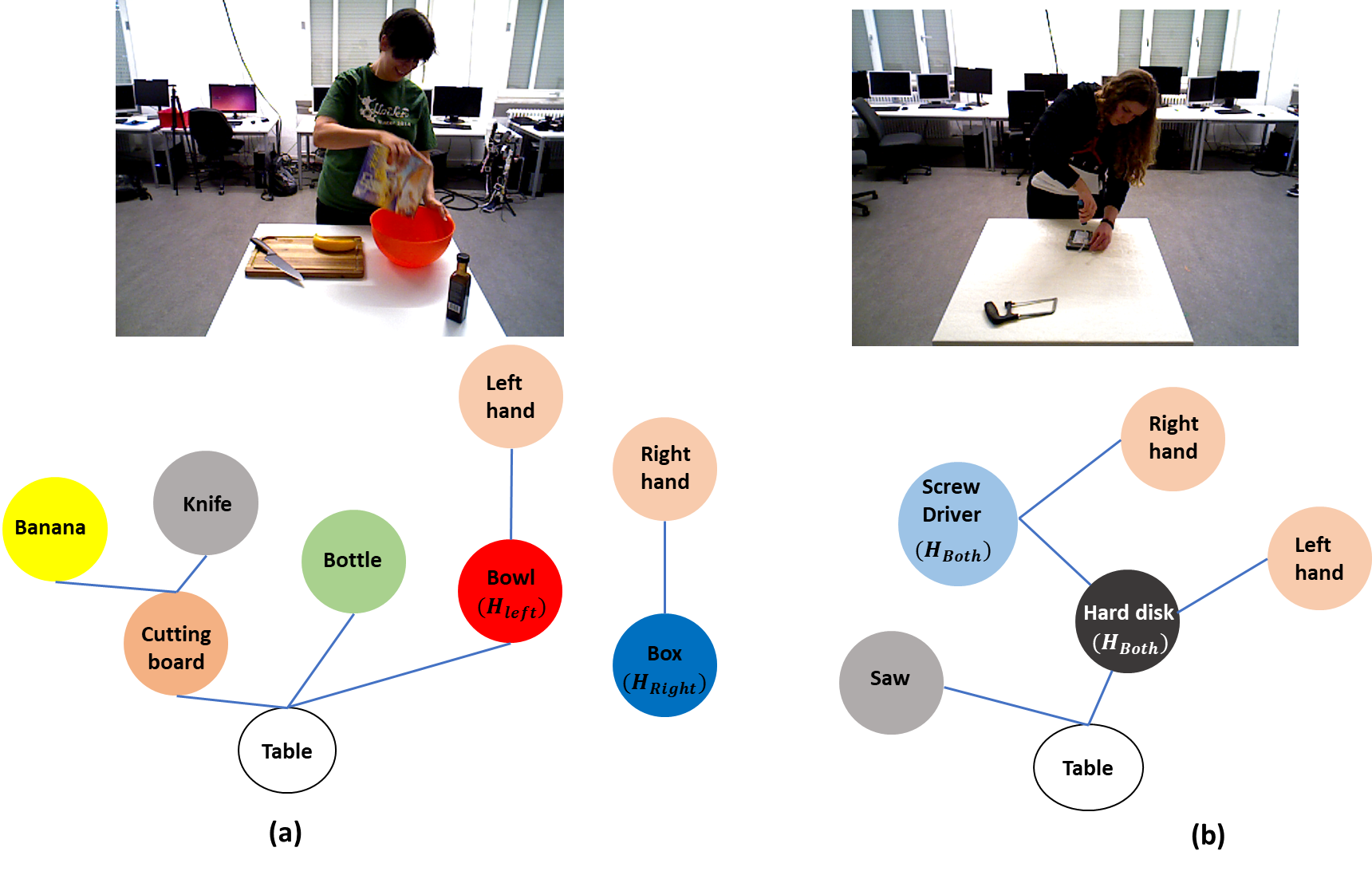}
\caption{An example of two scenes in the KIT bimanual action dataset \cite{dreher2020learning} and their corresponding contact graphs }
 \label{Fig.CG}
\end{figure}

In order to accurately compute bimanual action types, it is essential to determine which hands are involved in each bimanual interaction. Inspired by the work in \cite{krebs2022bimanual} to achieve this, we create a \textbf{contact graph} which is a binary adjacency matrix representation of the hand-object contacts in the scene graph. The contact graph is computed for each frame by examining the spatial relationships between objects and hands in the scene graph, and determining whether any of the hands is in contact with any of the objects \cite{aksoy2011learning}. Each entry in the contact graph indicates whether or not there is a contact between the corresponding nodes, where the rows and columns correspond to the nodes in the scene graph (i.e., hands and objects). A sample of a video frame and its corresponding contact graph is shown  Figure \ref{Fig.CG}. In the left scene the right hand is touching a box while the left hand is touching a bowl, therefore the box and the bowl are in the right hand group $ (H_{right}) $ and the left hand group $ (H_{left}) $, respectively. In this frame, the cutting board, knife, banana and bottle do not touch a hand (directly or indirectly) and therefore have no hand group (please notice the table as the main support surface is not considered as a contact).

The role of the contact graph in our framework is to provide a way to identify which hands are involved in each bimanual interaction. To achieve this, we compute hand groups based on the connectivity between hands in the contact graph. If two hands are connected in the contact graph (i.e., there is a path of one more more connecting edges between them), they are considered to belong to the same hand group. Hand groups are identified separately for each frame, so the hand group information is a sequence of hand group labels, one label for each frame. Let $\mathcal{G}_c = (\mathcal{V}_c, \mathcal{E}_c)$ represent the contact graph for a given frame, where $\mathcal{V}_c$ is the set of nodes representing hands and objects, and $\mathcal{E}_c$ is the set of edges indicating contacts between nodes. Suppose $\mathcal{H}$ is defined as the set of nodes representing hands, and $h_i$ and $h_j$ are two hand nodes. Then, $h_i$ and $h_j$ belong to the same hand group if and only if there exists a path between them in this graph. In the right scene of Figure \ref{Fig.CG}, the right hand is touching a screw driver and the left hand is touching a hard disk, while the screwdriver and the hard disk have contact with each other, therefore there is a path between two hands and the screwdriver and the hard disk belong to both hand groups. In this scene the saw has no direct or indirect contact with the hands and therefore it belongs to no hand group. More details about these rule-based process-computation steps were discussed in \cite{krebs2022bimanual}. 

The hand group information is then used in section \ref{Taxonomy} to compute bimanual action types based on the position and velocity information of the hands. Without the hand group information, we would not be able to distinguish between bimanual interactions where both hands are involved and those where only one hand is involved. This information is important in our framework since it helps us to accurately classify the different types of bimanual actions that can occur.


\subsection{Bimanual Action Type Computation}
\label{Taxonomy}
\subsubsection{Concept}
Bimanual type computation is an essential step for producing detailed and informative descriptions of bimanual actions. It also allows us to better understand the nature of the actions and how they are performed which can be especially useful in domains such as cooking and sports.

Krebs  et al. \cite{krebs2022bimanual} proposed a taxonomy that characterizes bimanual actions based on the symmetry or asymmetry, the coordination or independence of hand movements, and the dominance of one hand over the other. This taxonomy considers the position and velocity information of each hand in order to categorize bimanual actions into one of four categories: \textbf{symmetric coordinated}, \textbf{asymmetric coordinated}, \textbf{symmetric uncoordinated}, and \textbf{asymmetric uncoordinated}. In symmetric coordinated actions, both hands move together in a coordinated manner, such as clapping or playing piano. In asymmetric coordinated actions, one hand is dominant and the other is subordinate, such as when one hand leads while the other follows during writing or playing a musical instrument. In symmetric uncoordinated actions, both hands move independently but in a similar manner, such as when waving both hands. In asymmetric uncoordinated actions, both hands move independently and in different ways, such as when using a knife and fork to eat. The details of this taxonomy and its measurements were discussed in \cite{krebs2022bimanual}.

However, in order to better suit our needs, we have extended the above taxonomy to include a subcategory for \textbf{hand spatial relationships}, which includes subcategories for \textbf{close-hand}, \textbf{crossed-hand}, and \textbf{stacked-hand} actions, under the existing category of symmetry and asymmetry. We have also introduced a new category for \textbf{level of precision}, which describes the degree of precision required in performing the bimanual actions. This new category provides a finer-grained description of the actions and can help to differentiate between actions that require high or low levels of precision. These additions to the taxonomy allow for a more detailed and comprehensive description of bimanual actions, which can lead to more accurate and informative video descriptions. Figure~\ref{Fig.bimanual} represents our three newly defined hand spatial relations in a bimanual action plus the output of involving different levels of precision in chopping a carrot which shows the importance of considering precision details in our recognition and description framework.

\begin{figure}[!h]
\centering
\includegraphics[width=0.98\textwidth]{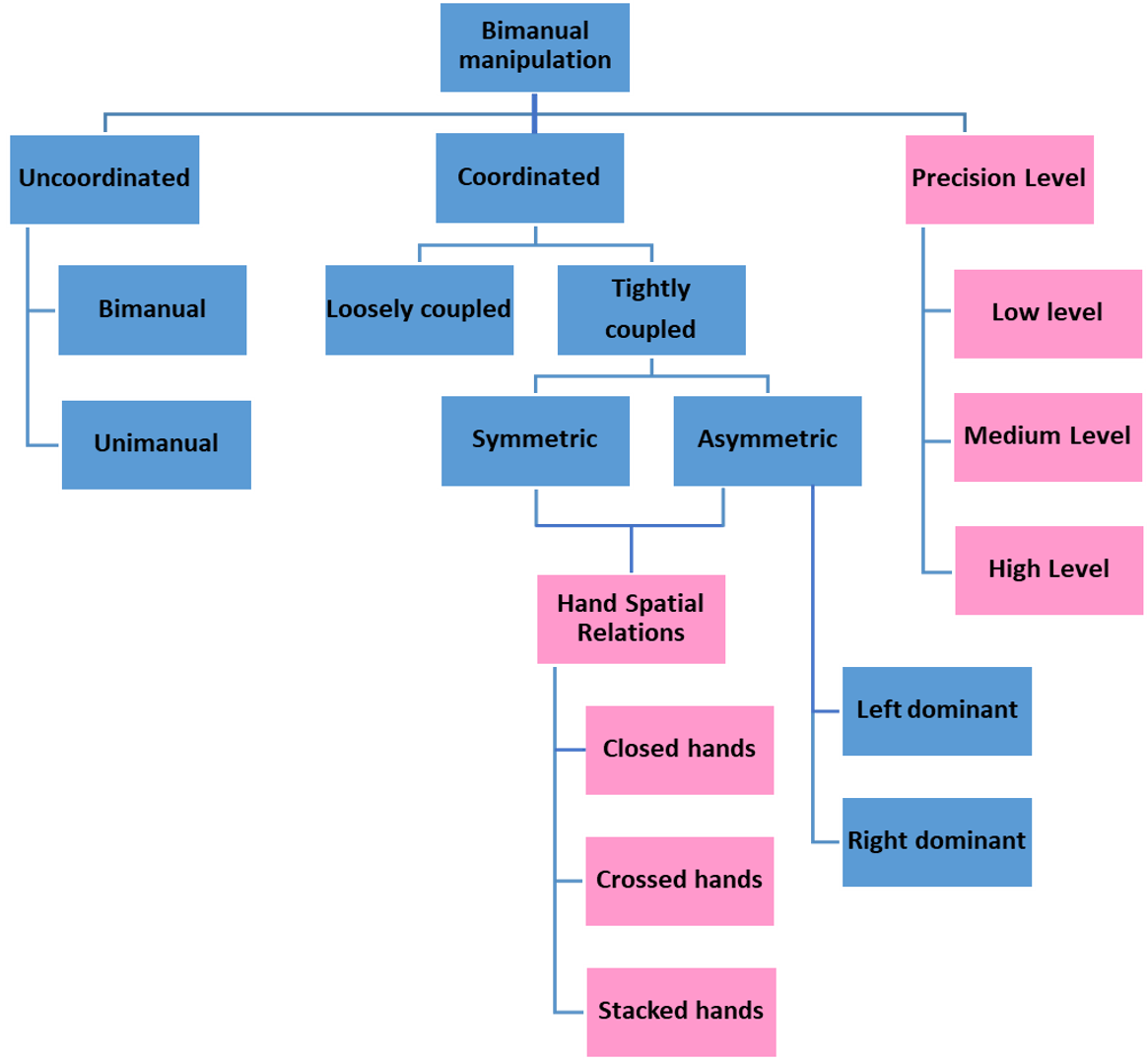}
\caption{Our modified bimanual action taxonomy inspired by the proposed taxonomy in \cite{krebs2022bimanual}}
 \label{Fig.taxonomy}
\end{figure}

The modified bimanual taxonomy we apply is depicted in Figure \ref{Fig.taxonomy}. In this taxonomy the blue cells were based on the Krebs et al. taxonomy \cite{krebs2022bimanual} while the pink cells depict the items in our modified taxonomy. 

Details how to determine had groups are found in the Supplementary Material and below (Results) we show a comparison of using different levels in the bimanual action taxonomy.

\subsection{Feature Aggregation}
\label{aggregation}

For GAT levels 1 and 2 (see e.g. Fig.~\ref{Fig.TCN}) we can directly use the feature vectors $\vec{g}$ from the TCN. For the third level, we concatenate the output of the TCN at level 3 (bimanual action) with a one-hot encoded vector representing the bimanual type. For simplicity we do not change notation and continue to use $\vec{g}$ and $\mathbf{G}$ for the feature vectors and its matrix -- non-aggregated or aggregated -- used in the next steps.

\subsection{Hierarchical Action Classification}
\label{classification}

The feature vectors $\vec{g}$ obtained above are used as inputs to a hierarchical action classification module. The goal of this module is to classify the action performed in the input video at multiple levels of detail based on the outputs of the GAT layers.

For each GAT layer $t$ where $1 \leq t \leq L$ and each action category $j$ among $N$ broad action categories, the feature matrix $\mathbf{G}^{(t,j)}$ is generated. These feature matrices encode different levels of information and granularity for action recognition in the $t$-th GAT layer (see the supplementary material for a better understanding).

In the context of $N$ action categories, each containing $M$ sublevels and each sublevel including $O$ items, the hierarchical action classification process is as follows. For each action category $j$, sublevel $k$, and item $o$, the feature matrix $\mathbf{G}^{(t,j,k,o)}$ is obtained from the GAT output at layer $t$.

Each feature matrix $\mathbf{G}^{(t,j,k,o)}$ undergoes a series of fully connected layers followed by a softmax function. The outcome is the action probability distribution $P^{(t,j,k,o)}$ corresponding to the $t$-th GAT layer, action category $j$, sublevel $k$, and item $o$. See Supplementary Material for details of the learning process.

\begin{figure}[!h]
\centering
\includegraphics[width=0.98\textwidth]{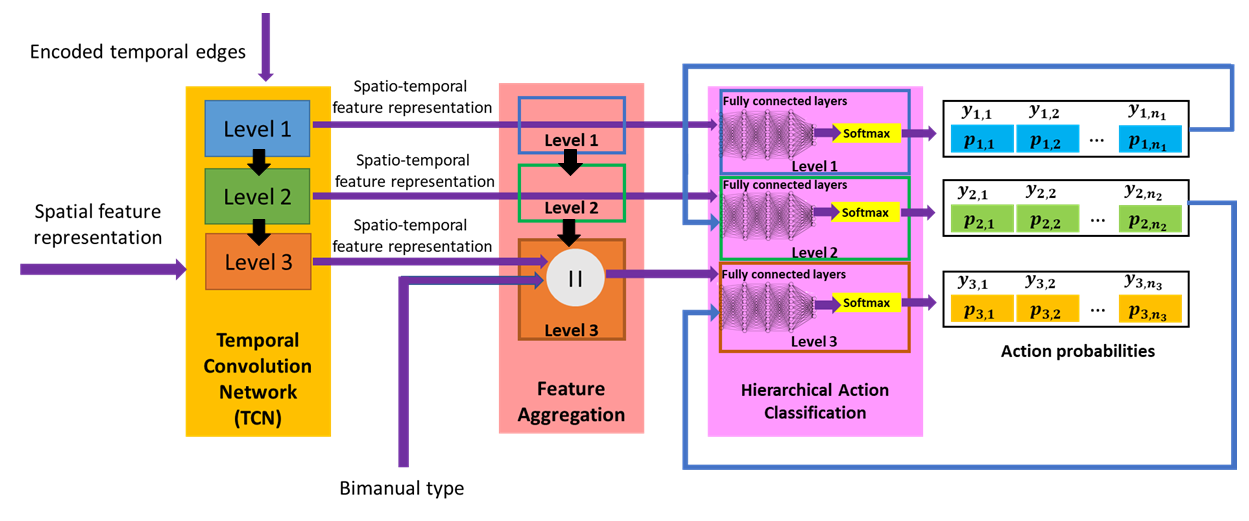}
\caption{Overview of the hierarchical action classification module. The feature vectors obtained from the feature aggregation step (Section~\ref{aggregation}) are used as inputs to a series of fully connected layers followed by a softmax function to obtain the action probability distribution at each level of the hierarchy. The predicted action labels at each level can be combined to form a complete action label for the entire video. In this figure, $y_{i,j}$ and $p_{i,j}$ represent the possible broad action categories and their corresponding probabilities at each level. Here for the simplicity we have only shown the broad action categories.}
 \label{Fig.classification}
\end{figure}

The predicted action label for the $t$-th GAT layer, action category $j$, sublevel $k$, and item $o$ is represented as $\hat{y}^{(t,j,k,o)} = \arg\max_{j} (P^{(t,j,k,o)})$. The classification process is illustrated in Figure \ref{Fig.classification}.

\subsection{Post-processing: Decision Making}
\label{post}
After obtaining the probability vectors for hierarchical action classification from (section \ref{classification}), the final decision for each level is made by choosing the action class with the highest predicted probability. However, due to the noisy nature of the prediction results, the output may contain sudden spikes or drops in the predicted probabilities, which can lead to inaccurate classification results. To address this issue, we apply a moving average filter, with a window size of $W$ frames, to smooth the predicted probability curves and improve the overall performance of the system. This filter is applied separately to each probability curve for each level, and the resulting smoothed curves are used to make the final decision.

Let $P^{(k)}_{j}(t)$ be the probability of action class $j$ at time $t$ for level $k$ and let $\hat{P}^{(k)}_{j}(t)$ be the corresponding smoothed probability. The moving average filter is defined as:

\begin{equation}
\hat{P}^{(k)}_{j}(t) = \frac{1}{W}\sum_{t-\lfloor W/2\rfloor}^{t+\lfloor W/2\rfloor} P^{(k)}_{j}(t)
\end{equation}

where time $t$ is defined as frame-steps with $\lfloor W/2\rfloor$ denoting the largest integer frame-number less than or equal to $W/2$, were we use $W=5$.

The smoothed probabilities $\hat{P}^{(k)}_{j}(t)$ are then used to make the final decision for each level, as follows:

\begin{equation}
\hat{y}^{(k)}(t) = \arg\max_j \hat{P}^{(k)}_{j}(t)
\end{equation}

where $\hat{y}^{(k)}(t)$ is the predicted action label at time $t$ and level $k$.

\begin{figure}[!h]
\centering
\includegraphics[width=0.99\textwidth]{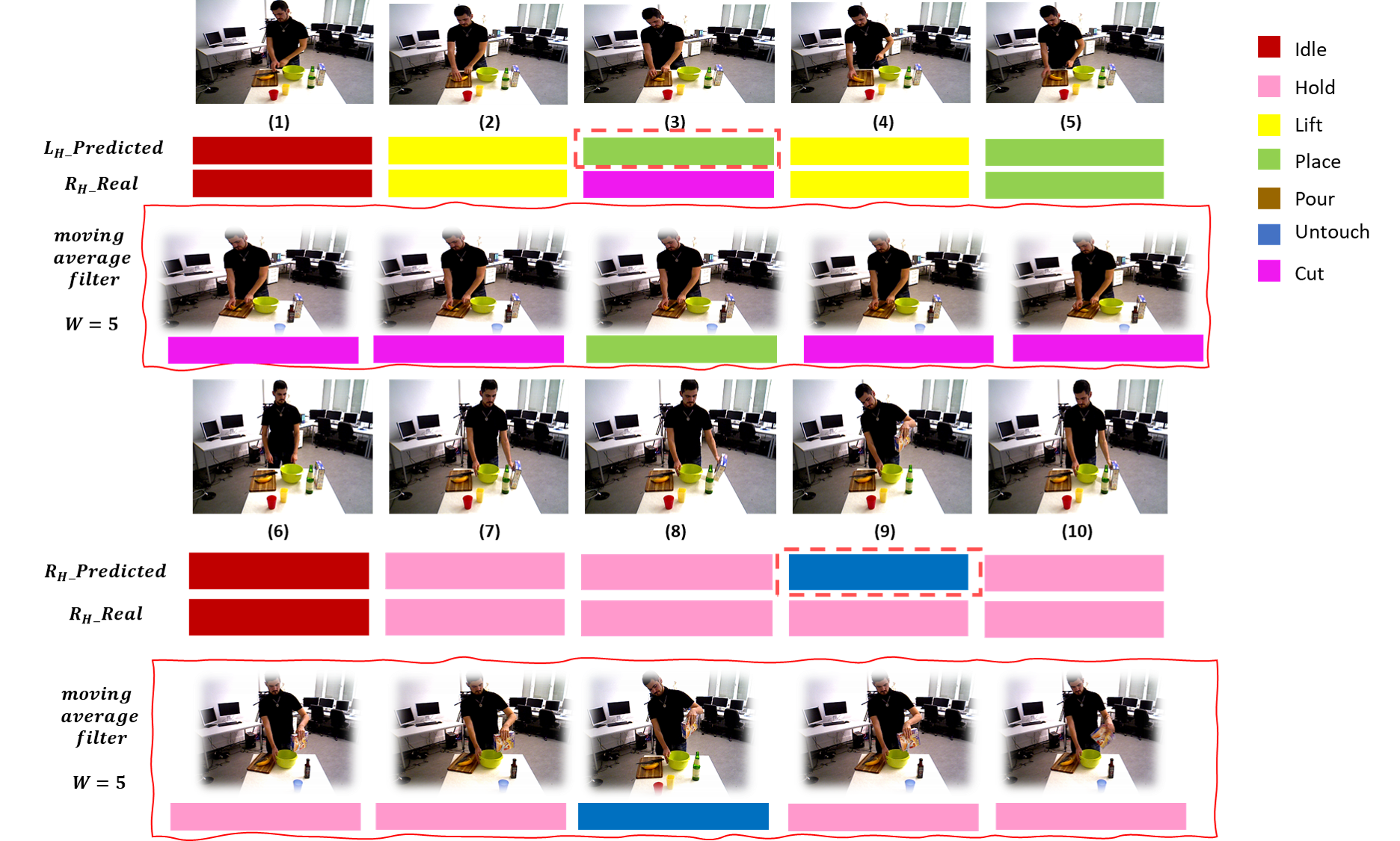}
\caption{The incorrect detection of the left-hand action in frame 3 and the incorrect detection of the right-hand action in frame 9 were corrected by applying a moving average filter with a window size of 5. The frames were selected from the KIT bimanual action dataset \cite{dreher2020learning}. 
}
 \label{Fig:postprocessing}
\end{figure}

To demonstrate the importance of post-processing and the use of smoothing techniques, we created a figure showing a man cutting a banana and pouring something into a bowl sequentially (Figure \ref{Fig:postprocessing}). During the cutting process, the action of his left hand was wrongly detected as ``place" instead of ``cut" in one frame, and in another frame, the action of his right hand was wrongly detected as ``untouch" instead of ``hold". However, by applying a moving average filter with a window size of 5, these mistakes were corrected. 

\subsection{Description Generation}
\label{description}

In recent years, pre-trained language models, such as GPT-2, \cite{schramowski2022large}\cite{gira2022debiasing}\cite{zhou2020evaluating}\cite{yu2023fusing} have shown promising results in various natural language processing tasks. 
In this section, we discuss the use of a pre-trained GPT-2 model for generating bimanual action descriptions. 

\begin{figure}[!h]
\centering
\includegraphics[width=0.99\textwidth]{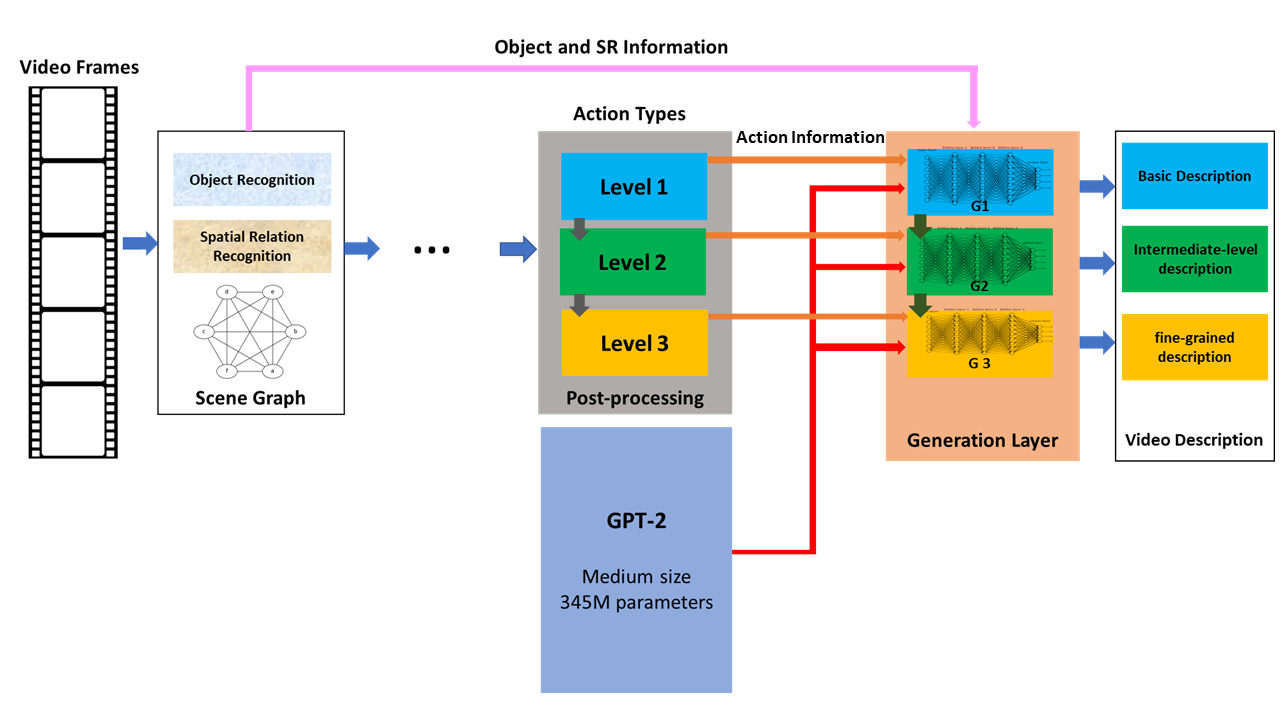}
\caption{GPT-2 Based architecture for bimanual action description generation}
 \label{fig:Generation}
\end{figure}

The GPT-2 model has a relatively simple architecture compared to other models such as GPT-3 \cite{brown2020language}, BERT \cite{devlin2018bert}, and T5 \cite{raffel2019exploring}, which makes it easier to fine-tune and adapt to our task. We use the GPT-2 medium model, which has 345 million parameters and has been trained on a diverse range of text data, including web pages, books, and articles. Figure~\ref{fig:Generation} shows a schematic of how GPT-2 enters our system.




However, to adapt the pre-trained GPT-2 model to the task of generating bimanual action descriptions, we fine-tune the model on our annotated dataset. The fine-tuning process involves updating the weights of the pre-trained model to better capture the patterns and relationships in our dataset. The fine-tuning process consists of the following steps (See Supplementary Material for implementation details):

\begin{itemize}

\item Tokenization of the input data.

\item Vectorization of these tokens to create numerical representations.

\item Using a sliding windows to handle longer input sentences.

\item Designing a specific model architecture adding generation layers responsible for generating the action descriptions.

\item Performing weight adaptation of GPT-2 with a cross-entropy loss function and by using the Adam optimizer.

\end{itemize}

The fine-tuning process is complete when the model has achieved satisfactory performance on our annotated dataset for bimanual action description generation and can generate accurate and coherent descriptions for new input examples in the context of our specific task.

\begin{figure}[!h]
\centering
\includegraphics[width=0.98\textwidth]{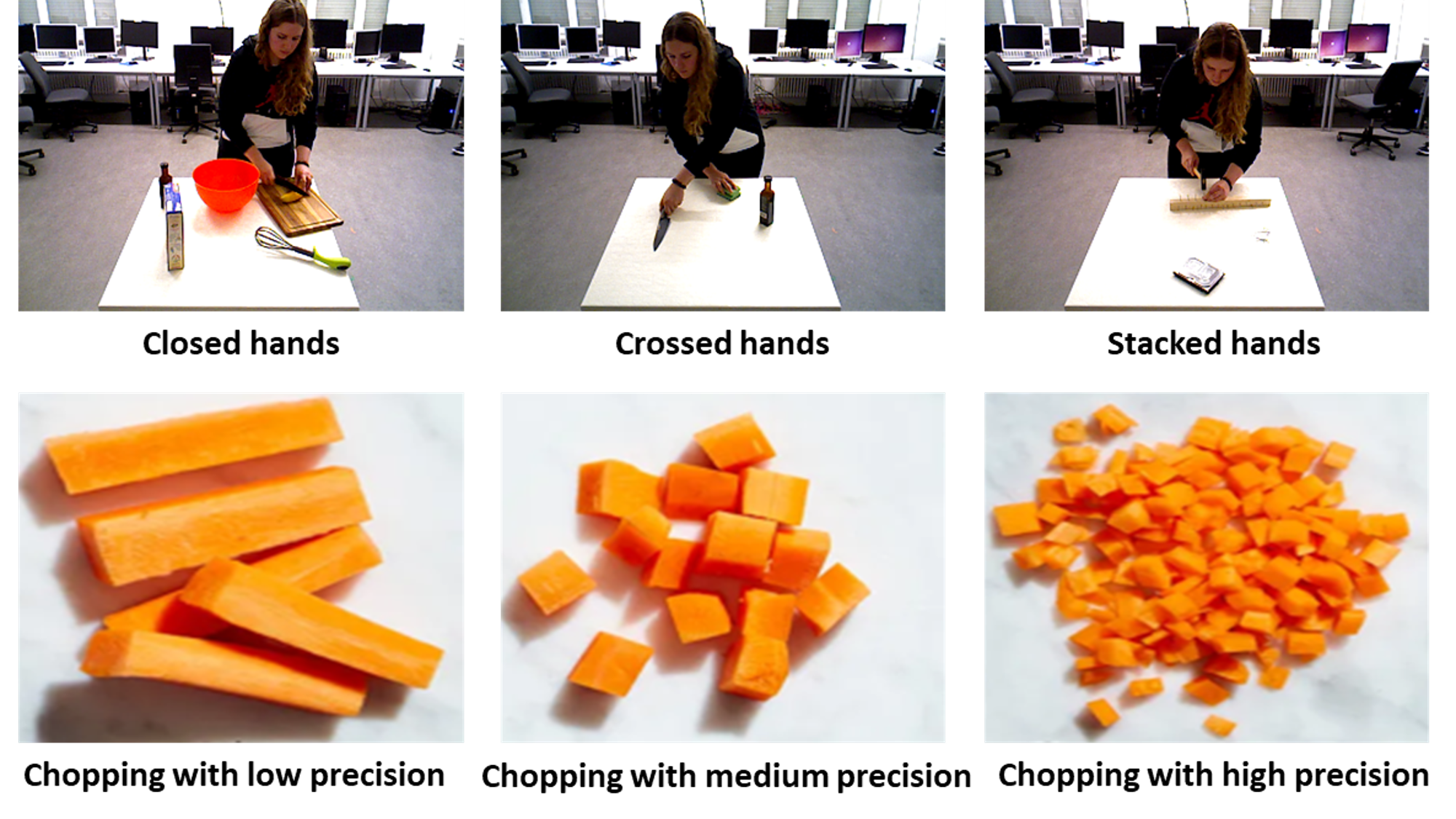}
\caption{\textbf{Top row:} Crossed hands, close hands and stacked hands in wiping, cutting and hamming actions from the KIT bimanual action dataset \cite{dreher2020learning}.\\ \textbf{Bottom row:} Carrot pieces after chopping with low, medium and high levels of precision}
 \label{Fig.bimanual}
\end{figure}

\section{Results and Discussion}
\label{results}

\subsection{Qualitative Results}
In Figure~\ref{Fig.bimanual} we first demonstrate the effectiveness of our modified taxonomy. For this we do a comparison. 

\begin{itemize}

\item \textbf{Without involving bimanual type:} ``A person is performing a chopping action on a cutting board."

\item \textbf{With involving Krebs taxonomy \cite{krebs2022bimanual}:} ``A person is performing an asymmetric and coordinated chopping action with a knife in his/her right hand and a cutting board in his/her left hand."

\item \textbf{With involving our modified taxonomy:} ``A person is performing an asymmetric and coordinated chopping action with a knife in his/her right hand and a cutting board in his/her left hand, while maintaining a stacked-hand spatial relationship and a high level of precision in cutting the vegetables into small pieces."

\end{itemize}

As you see with the modified taxonomy we can better capture the nuances and complexities of the action, which can be important in certain contexts. For example, in a surgical setting where delicate and precise bimanual actions are required, having information about the hand spatial relationship and level of precision can be important for describing the success of the procedure.

While computing this added information may require some additional computational load, the benefits of having more detailed and precise descriptions can be significant in certain contexts, and may be worth the additional effort.

The general qualitative performance is shown in the samples of Figure \ref{Fig.three_levels}. They show a representative video frame for context and depicts the three levels of our hierarchical framework for generating video descriptions, along with the amount of detail included at each level. 

\subsection{Quantifications}
Our evaluation is structured into several subsections, each focusing on specific aspects of our research. We begin by assessing the recognition of object-action relations, followed by an evaluation of manipulation action recognition, bimanual action recognition, and description generation. Additionally, we provide a cross-domain evaluation to gauge the generalization capabilities of our model across different datasets. Furthermore, we compare our results against state-of-the-art methods to underscore the benefits of our approach. 

Note that in the Supplement we provide details about the different data sets, describing why they are useful for our tasks, in the main text we only name them.

\subsubsubsection{Object-Action Relation Evaluation}
\label{subsec:object-action-evaluation}

In this subsection, we evaluate the performance of our proposed method in recognizing object-action relations.\\

\noindent
\underline{\bf Datasets:} MS-COCO \cite{lin2014microsoft} and SomethingSomething \cite{goyal2017something}. For our experiments, we select the subset of both datasets that specifically focuses on manipulation actions. This subset includes relevant object categories and action labels associated with manipulation actions. By using this subset, we can concentrate our evaluation on the specific domain of object-action relations and obtain more targeted insights into the performance of our algorithm.




\noindent
\underline{\bf Performance Comparison:} To assess the performance of our method in learning object-action relations, we compare it with the following selected state-of-the-art (SOA) approaches: \cite{ji2020action}, \cite{wang2020learning}, \cite{kim2021hotr}, and \cite{dreher2020learning}. These approaches have made significant contributions to the field of human-object interaction and action recognition. 

The performance comparison is summarized in Table \ref{table:performance}, which presents the results in terms of F1-score, recall, and precision for each approach on the two datasets. The F1-score provides a measure of the overall accuracy, while recall and precision offer insights into the completeness and correctness of object-action relation predictions.

\begin{table}[htbp]
\caption{`Performance Comparison of Object-Action Relation Learning between Our Method and State-of-the-Art Approaches on MS-COCO and Something-Something Datasets}
\label{table:performance}
\begin{tabular}{|l|c|c|c|c|c|c|}
\hline
\multirow{2}{*}{Approach} & \multicolumn{3}{c|}{MS-COCO} & \multicolumn{3}{c|}{Something-Something} \\
\cline{2-7}
& F1-Score & Recall & Precision & F1-Score & Recall & Precision \\
\hline
Our Method & \textbf{0.83} & \textbf{0.86} & \textbf{0.81} & \textbf{0.78} & \textbf{0.80} & \textbf{0.76} \\
\cite{ji2020action} & 0.70 & 0.77 & 0.64 & 0.67 & 0.75 & 0.60 \\
\cite{wang2020learning} & 0.63 & 0.72 & 0.56 & 0.61 & 0.70 & 0.54 \\
\cite{kim2021hotr} & 0.66 & 0.61 & 0.72 & 0.64 & 0.59 & 0.70 \\
\cite{dreher2020learning} & 0.69 & 0.64 & 0.75 & 0.66 & 0.61 & 0.72 \\
\hline
\end{tabular}
\end{table}




The results of our method demonstrate its strong performance in both precision and recall, and also the resulting high F1-Score, indicating that it achieves a low false positive rate and effectively captures a large proportion of the actual positive instances. In contrast, some of the compared methods show a clear distinction between precision and recall, with one metric performing better than the other (their corresponding recall or precision is high but their F1-Score is not). This suggests that these methods may excel in either accurately identifying positive instances or capturing a comprehensive set of relevant instances, but do not achieve both simultaneously.

\noindent
\underline{\bf Discussion:} In the field of object-action relations, several notable approaches have been proposed, including those by Ji et al. \cite{ji2020action}, Wang et al. \cite{wang2020learning}, Kim et al. \cite{kim2021hotr}, and Dreher et al. \cite{dreher2020learning}.

Ji et al. \cite{ji2020action} present a framework that represents actions as compositions of spatio-temporal scene graphs. This approach focuses on capturing the structural relationships between objects and actions within a scene. Similarly, Wang et al. \cite{wang2020learning} propose an approach that models spatial relationships between interaction points to detect human-object interactions. Both methods contribute to the understanding of object-action relations but do not explicitly emphasize the detailed learning and representation of such relations.

Kim et al. \cite{kim2021hotr} introduce an end-to-end framework that utilizes transformers for human-object interaction detection. This approach benefits from the ability of transformers to capture global context and long-range dependencies. However, it does not explicitly focus on the fine-grained learning of object-action relations.

By contrast, our method stands out by placing a specific emphasis on learning and representing detailed object-action relations through the utilization of graph networks and advanced relational modeling techniques. By capturing the complex interplay between objects and actions more effectively, our method achieves superior results in object-action relation learning. It excels at capturing fine-grained dependencies and interactions between objects and actions, resulting in a more comprehensive understanding of object-action dynamics. 

Dreher et al. \cite{dreher2020learning} on the other hand directly focus on understanding of object-action relations but their work has limitations related to dataset dependence, scalability, sensitivity to variations, handling ambiguous relationships, and capturing temporal dynamics. Their approach does not explicitly address the temporal aspects of object-action relations. While they focus on learning relationships from demonstrations, the temporal dynamics and sequential patterns of object-action interactions may not be fully captured or explicitly modeled in their method but our approach considers the temporal dynamics of object-action relations. By explicitly modeling and incorporating sequential patterns and temporal dependencies, we capture the logical sequences of actions and their relationships over time. This temporal awareness contributes to a more comprehensive understanding of object-action dynamics and improves the overall accuracy and robustness of our approach.

Moreover, our method exhibits scalability and efficiency in handling larger datasets and complex scenarios. We have designed our approach to address computational complexity and memory requirements, ensuring its feasibility and effectiveness even with increasing problem scales.

\subsubsection{Manipulation Action Recognition Evaluation}
\label{mani_res}
In this section, we evaluate the performance of our proposed method for manipulation action recognition on both, 2D and 3D, datasets. The evaluation includes a comprehensive analysis of performance metrics and comparisons with state-of-the-art methods.\\

\noindent
\underline{\bf 2D and 3D Datasets:} To assess the effectiveness of our proposed method for manipulation action recognition in 2D, we selected several widely used datasets known for their diverse collection of manipulation actions. We used the following 2D datasets: Youcook2  \cite{zhou2018automatic}, Charades-STA  \cite{gao2017tall}, ActivityNet Captions  \cite{krishna2017dense}, TACoS, \cite{regneri2013grounding}.
To evaluate the performance of our proposed method in a 3D context, we selected datasets that provide realistic and immersive environments for manipulation action recognition which are: 
KIT Bimanual Action Dataset  \cite{dreher2020learning}, YCB-Video Dataset \cite{xiang2017posecnn}, G3D Dataset \cite{bloom2012g3d}.

By utilizing these 3D datasets, we can evaluate the performance of our proposed method in realistic 3D environments, where precise object segmentations and depth information enhance the accuracy and robustness of manipulation action recognition.

The inclusion of both 2D and 3D datasets in our evaluation ensures a comprehensive assessment of our proposed method's capabilities in recognizing and understanding manipulation actions across different modalities and contexts.

\noindent
\underline{\bf Performance Comparison:} In this section, we compare the performance of our proposed approach for manipulation action recognition with three state-of-the-art papers: \cite{ji2020action}, \cite{wang2020learning}, and \cite{wen2023hierarchical}. These papers represent the state-of-the-art in manipulation action recognition.

The selected papers are comparable to our work as they address similar research objectives and propose novel methodologies for manipulation action recognition. They provide solutions for capturing spatial and temporal dependencies, recognizing human-object interactions, and using egocentric RGB videos for accurate action recognition.

To compare our approach with these papers, we present a table summarizing the key aspects and performance metrics on the selected datasets (Table \ref{tab:comparison}).

\begin{table}[ht]
\centering
\caption{Performance Comparison of Manipulation Action Recognition Approaches.}
\label{tab:comparison}
\resizebox{\textwidth}{!}{%
\begin{tabular}{|c|c|c|c|}
\hline
\textbf{Approach} & \textbf{Dataset} & \textbf{Accuracy} & \textbf{Precision} \\
\hline
\multirow{7}{*}{Our Method} & Youcook2 & \textbf{0.85} & \textbf{0.79} \\
& Charades-STA & \textbf{0.83} & 0.76 \\
& ActivityNet Captions & \textbf{0.89} & \textbf{0.92} \\
& TACoS & \textbf{0.80} & \textbf{0.89} \\
& Kit Bimanual Action Dataset & \textbf{0.91} & \textbf{0.88} \\
& YCB-Video Dataset & \textbf{0.84} & 0.79 \\
& G3D Dataset & \textbf{0.89} & \textbf{0.84} \\
\hline
\multirow{7}{*}{\cite{ji2020action}} & Youcook2 & 0.82 & 0.77 \\
& Charades-STA & 0.75 & \textbf{0.82} \\
& ActivityNet Captions & 0.82 & 0.78 \\
& TACoS & 0.76 & 0.72 \\
& Kit Bimanual Action Dataset & 0.73 & 0.69 \\
& YCB-Video Dataset & 0.64 & 0.68 \\
& G3D Dataset & 0.80 & 0.76 \\
\hline
\multirow{7}{*}{\cite{wang2020learning}} & Youcook2 & 0.66 & 0.75 \\
& Charades-STA & 0.72 & 0.68 \\
& ActivityNet Captions & 0.68 & 0.65 \\
& TACoS & 0.73 & 0.69 \\
& Kit Bimanual Action Dataset & 0.78 & 0.74 \\
& YCB-Video Dataset & 0.69 & 0.66 \\
& G3D Dataset & 0.70 & 0.63 \\
\hline
\multirow{3}{*}{\cite{wen2023hierarchical}} & Kit Bimanual Action Dataset & 0.81 & 0.78 \\
& YCB-Video Dataset & \textbf{0.84} & \textbf{0.85} \\
& G3D Dataset & 0.86 & 0.82 \\
\hline
\end{tabular}%
}
\end{table}

The comparison table provides a quantitative evaluation of the performance of each approach on the specified datasets in recognizing manipulation actions.

In the extensive evaluation conducted across a variety of datasets, our method consistently showcased exemplary performance, outperforming the competing state-of-the-art approaches in most instances. Nonetheless, there were specific scenarios where our method did not clinch the top spot. Particularly, in the YCB-Video Dataset, our approach, despite demonstrating strong results, was slightly edged out by the method introduced by Wen et al. \cite{wen2023hierarchical}, which secured the highest precision scores \(0.85\), while having the same accuracy with our method \(0.84\). This variance can be attributed to the nuanced complexities and unique challenges presented by the dataset, which may have been more effectively addressed by the hierarchical attention mechanisms employed by Wen et al. In addition, while our method exhibited leading accuracy on the Charades-STA dataset, it was narrowly outperformed in precision, with the method by Ji et al. \cite{ji2020action} recording the highest precision of \(0.82\). This divergence in performance might be due to the adaptability of Ji et al.'s approach to diverse and dynamic environments inherent in the dataset, allowing for more accurate identification and tracking of multiple objects and their interactions over time. These instances not only underscore the multifaceted challenges posed by different datasets but also serve as invaluable learning points, guiding future refinements and advancements in our methodology.

\noindent
\underline{\bf Discussion:} Wen et al. \cite{wen2023hierarchical} introduced a hierarchical temporal transformer for 3D hand pose estimation and action recognition from egocentric RGB videos. Their method utilizes hierarchical attention mechanisms to capture fine-grained temporal dynamics in hand actions. However, when compared to our approach, their method does not demonstrate comparable performance in manipulation action recognition tasks. This can be attributed to several factors, including limitations in the modeling of spatial and temporal relationships, potential challenges in handling complex object interactions, and potential difficulties in capturing nuanced hand-object dynamics.

Ji et al. \cite{ji2020action} and Wang et al. \cite{wang2020learning} as the other comparison candidates hd partially been discussed in Section \ref{subsec:object-action-evaluation} already. Ji et al.  involve understanding both, the objects present in a scene (spatial aspect) and how these objects interact or change over time (temporal aspect). This creates challenges in terms of the complexity of data representation, scalability, and computational efficiency. The approach struggles with accurately identifying and tracking multiple objects and their interactions over time, especially in cluttered or dynamically changing environments. Moreover, a common challenge in Human-Object Interaction (HOI) detection in Wang et al.'s method is the diversity and complexity of possible interactions between humans and objects. Using interaction points for HOI detection does not capture the complete semantics of interactions, particularly for complex or multi-stage actions.

In comparison, our approach overcomes these limitations by combining the power of graph-based modeling and attention mechanisms. By explicitly representing hand-object relationships as a scene graph and incorporating hierarchical attention mechanisms, our method captures the intricate spatial and temporal dependencies involved in manipulation actions. This enables us to achieve more accurate and robust recognition of manipulation actions, even in complex and challenging scenarios. Furthermore, our approach takes into account the semantic and contextual information of hand-object interactions, leading to a more comprehensive understanding of manipulation actions and improved recognition performance.

\subsubsection{Bimanual Action Recognition Evaluation}
\noindent
\underline{\bf Datasets:} In order to evaluate the performance of our proposed bimanual action recognition approach, we selected the following datasets: 
KIT Bimanual Action Dataset \cite{dreher2020learning}, YCB-Video Dataset \cite{xiang2017posecnn},G3D Dataset \cite{bloom2012g3d}, CMU Fine Manipulation Dataset \cite{calli2017yale}. These datasets provide a diverse range of bimanual manipulation actions and enable us to evaluate the performance of our approach in recognizing and understanding bimanual action types, object interactions, and precision levels.



\noindent
\underline{\bf Performance Comparison:} In this subsection, we present a detailed performance comparison of our proposed bimanual action recognition approach in two key aspects: bimanual action recognition and bimanual taxonomy identification. We compare the performance of our approach with two state-of-the-art papers: Dreher et al. \cite{dreher2020learning} and Andrade et al. \cite{andrade2022human}. 

Table \ref{tab:action_recognition} summarizes the performance comparison for bimanual action recognition across the evaluated datasets.

\begin{table}[ht]
\centering
\caption{Performance Comparison for Action Recognition}
\label{tab:action_recognition}
\begin{tabular}{|c|c|c|c|}
\hline
\textbf{Approach} & \textbf{Dataset} & \textbf{Accuracy} & \textbf{Precision} \\
\hline
\multirow{4}{*}{Our Method} & KIT Bimanual Action & \textbf{0.87} & \textbf{0.85} \\
& YCB-Video & \textbf{0.80} & \textbf{0.82} \\
& G3D & \textbf{0.86} & \textbf{0.82} \\
& CMU Fine Manipulation & \textbf{0.77} & \textbf{0.83} \\
\hline
\multirow{4}{*}{\cite{dreher2020learning}} & KIT Bimanual Action & 0.83 & 0.80 \\
& YCB-Video & 0.81 & 0.78 \\
& G3D & 0.83 & 0.76 \\ 
& CMU Fine Manipulation & 0.64 & 0.73 \\
\hline
\multirow{4}{*}{\cite{andrade2022human}} & KIT Bimanual Action & 0.80 & 0.76 \\
& YCB-Video & 0.78 & 0.66 \\
& G3D & 0.75 & 0.70 \\
& CMU Fine Manipulation & 0.71 & 0.67 \\
\hline
\end{tabular}
\end{table}

The performance comparison table elucidates the efficacy of our method across various datasets. Our method predominantly outperforms the other approaches, achieving the highest accuracy and precision in three out of the four datasets. Notably, for the KIT Bimanual Action, YCB-Video, and CMU Fine Manipulation datasets, our method exhibits superior performance, underscoring its robustness and versatility in action recognition tasks.

However, it is important to acknowledge that our method does not universally dominate in all scenarios. Specifically, in the G3D dataset, the approach proposed by Dreher et al. \cite{dreher2020learning} slightly surpasses ours, achieving an accuracy and precision of 0.86 and 0.82, respectively, compared to our 0.83 and 0.76. This divergence in performance could be attributed to the inherent characteristics and challenges posed by the G3D dataset, potentially favoring the techniques employed by Dreher et al.

Nevertheless, the overall superior performance of our method across diverse datasets underscores its adaptability and effectiveness in manipulation action recognition, demonstrating its potential as a valuable tool for advancing research in this domain.\\

\noindent
\underline{\bf Discussion:} Dreher et al. \cite{dreher2020learning} specifically concentrated on bimanual action recognition using graph neural networks. Their method demonstrated promising results in capturing object-action relations and temporal dependencies in bimanual scenarios. On the other hand, Andrade et al. \cite{andrade2022human} focused on general action recognition without explicitly targeting bimanual actions. While their approach did not directly address bimanual actions, their method incorporated spatial and temporal dependencies between objects and actions. This aspect is particularly relevant in understanding bimanual action scenarios where interactions between objects and hands play a crucial role.

Our approach was initially inspired by Dreher et al.'s method, as both approaches leverage scene graphs and graph neural networks (GNNs) for bimanual action recognition. However, there are key differences that contribute to the superior performance of our approach.

Firstly, the architecture of Dreher et al.'s method is simpler, consisting of an encoder, a core, and a decoder including a simple two-layer perceptron as update function and a sum function for node and edge aggregation. In contrast, our approach incorporates a more complex architecture with a hierarchical Graph Attention Network (GAT). The GAT's multi-head attention mechanism allows our model to capture more accurate and informative features, as it attends to different aspects of the scene graph and captures both local and global contextual information.

Furthermore, the treatment of spatial relations and temporal edges differs between the two approaches. In Dreher et al.'s method, spatial relations and temporal edges are treated as mutually exclusive. This means that they do not consider the interaction between spatial and temporal aspects of the bimanual actions. In contrast, our approach explicitly incorporates both spatial and temporal dependencies by considering temporal concatenation of scene graphs. This enables our model to capture the temporal evolution of object relationships and actions, providing a more comprehensive understanding of bimanual actions.

Additionally, Dreher et al. utilize one-hot vectors for encoding object and action labels, representing each category as a binary feature. In contrast, we leverage from learned embeddings for object and action labels, which capture more nuanced and continuous semantic representations. This allows our model to focus on objects that have a more significant contribution to the action progression, rather than assigning equal weights to all objects. As a result, our approach achieves greater efficiency and discriminative power in capturing relevant object-action relationships.

Furthermore, the comparison between our work and the method proposed by Andrade et al. reveals that our approach excels in terms of accuracy and precision, despite both methods incorporating the temporal dimension in activity recognition. One of the key strengths of our approach lies in the utilization of graph neural networks (GNNs), which allows us to capture complex spatial and temporal relationships. By representing interactions between objects as a graph, our model gains a deeper understanding of human activities. In contrast, Andrade et al.'s method relies on a temporal convolutional neural network (CNN) architecture, primarily focusing on capturing temporal patterns. While their approach demonstrates competence in activity recognition, our use of graph-based modeling enables more detailed analysis of object relationships and captures fine-grained temporal dynamics. Additionally, our approach incorporates a hierarchical attention mechanism, specifically Graph Attention Networks (GATs), which enhances precision and discriminative power in activity recognition. The hierarchical attention mechanism allows our model to selectively attend to relevant objects and temporal features, further improving the accuracy of activity recognition. In contrast, Andrade et al.'s method does not include such an attention mechanism, limiting their ability to capture fine-grained spatial and temporal dependencies.\\

\noindent
\underline{\bf Bimanual Action Types and Hand Spatial Relations: } In this section, we evaluate our enriched version of the bimanual action taxonomy compared to the taxonomy proposed by Krebs et al. \cite{krebs2022bimanual}, as illustrated in Fig.\ref{Fig.taxonomy}. While our method aligns with their approach in recognizing the blue cells in the taxonomy, we have introduced a novel method to accurately identify hand spatial relations and precision levels, denoted by the pink cell (Fig.\ref{Fig.taxonomy}). As no other taxonomy method exists in the literature for comparison, we focus on reporting the accuracy of our system across the datasets.

\begin{table}[ht]
\centering
\caption{Accuracy of Bimanual Action Type Determination (Blue Cell Items in Fig.\ref{Fig.taxonomy})}
\label{tab:bimanual_taxonomy_a}
\resizebox{\textwidth}{!}{%
\begin{tabular}{|c|c|c|c|c|}
\hline
\textbf{Dataset} & \textbf{Uncoordinated} & \textbf{Symmetric} & \textbf{Asymmetric} & \textbf{Dominant Hand} \\
\hline
KIT Bimanual Action & 0.87 & 0.92 & 0.89 & 0.88 \\
YCB-Video & 0.84 & 0.91 & 0.86 & 0.85 \\
G3D & 0.89 & 0.93 & 0.90 & 0.89 \\
CMU Fine Manipulation & 0.86 & 0.90 & 0.87 & 0.86 \\
\hline
\end{tabular}%
}
\end{table}

\begin{table}[ht]
\centering
\caption{Accuracy of Bimanual Action Type Determination (Pink Cell Items in Fig.\ref{Fig.taxonomy})}
\label{tab:bimanual_taxonomy_b}
\resizebox{\textwidth}{!}{%
\begin{tabular}{|c|c|c|c|c|c|}
\hline
\multirow{2}{*}{\textbf{Dataset}} & \multicolumn{3}{c|}{\textbf{Hand Spatial Relation}} & \multirow{2}{*}{\textbf{Precision Level}} \\
\cline{2-4}
 & \textbf{Close Hands} & \textbf{Crossed Hands} & \textbf{Stacked Hands} & \\
\hline
KIT Bimanual Action & 0.91 & 0.88 & 0.93 & 0.92 \\
YCB-Video & 0.88 & 0.87 & 0.90 & 0.89 \\
G3D & 0.90 & 0.89 & 0.92 & 0.91 \\
CMU Fine Manipulation & 0.89 & 0.86 & 0.91 & 0.88 \\
\hline
\end{tabular}%
}
\end{table}

As shown in Table \ref{tab:bimanual_taxonomy_a}, our system achieves notable accuracy in determining bimanual action types. The accuracy scores range from 0.84 to 0.93, indicating the robustness and effectiveness of the method. Specifically, our system consistently achieves accuracy levels above 0.85 for symmetric and asymmetric actions, as well as accuracy levels above 0.86 for uncoordinated actions. Moreover, the accuracy in determining the dominant hand ranges from 0.85 to 0.89, demonstrating the reliability of our approach.

Additionally, our method exhibits impressive precision in identifying hand spatial relations, as depicted in the second table of Table \ref{tab:bimanual_taxonomy_a}. The precision scores also range from 0.88 to 0.92 for finding hand spatial relations regarding each other.

While a direct comparison with existing taxonomy methods is not possible due to the absence of alternative approaches in the literature, our enriched taxonomy reports convincing accuracies across multiple datasets. These notable accuracy scores highlight the reliability of our approach, which holds significant potential for various applications, including video description generation. Moreover this way, robots can better understand and replicate human actions, leading to more efficient human-robot collaboration.

\subsubsection{Description Generation Evaluation}
\noindent
\underline{\bf Datasets: } For evaluating the performance of our description generation approach, we have selected a set of datasets that provide diverse and annotated video descriptions. These datasets allow us to assess the effectiveness of our method in generating accurate and informative descriptions of both 2D and 3D manipulation and bimanual actions. For 2D Datasets we used: Youcook2 Dataset \cite{zhou2018automatic}, TACoS Multi-Level Dataset \cite{regneri2013grounding}, ActivityNet Captions Dataset \cite{krishna2017dense}, Charades Dataset \cite{sigurdsson2016hollywood}.

For 3D Datasets we used the KIT Bimanual Action Dataset \cite{dreher2020learning}.

These datasets were chosen for their strong relevance to manipulation and bimanual actions, as well as their availability of natural language descriptions or annotations. 
Please note that although datasets like MSR-VTT \cite{xu2016msr} and MSVD \cite{chen2011collecting} are also commonly used in state-of-the-art research on video description but could not be used here because of their limitations concerning the level of annotation detail (e.g. no bimanual annotations).

\noindent
\underline{\bf Performance Comparison:} In this subsection, we compare the performance of our description generation approach with three state-of-the-art papers: Wang et al. (2018) \cite{wang2018reconstruction}, Seo et al. (2022) \cite{seo2022end}, and Lin et al. (2022) \cite{lin2022swinbert}. These papers represent significant contributions to the field of video captioning and provide valuable insights into different approaches and techniques.

To evaluate the performance, we compare our approach with the aforementioned papers using several commonly used evaluation metrics, including BLEU-4, METEOR, ROUGE-L, and CIDEr. These metrics provide a quantitative assessment of the quality and similarity of the generated descriptions compared to the ground truth annotations.

Table \ref{tab:performance_comparison} summarizes the performance of our approach and the three selected papers on the various datasets used in our evaluation.

\begin{table}[ht]
\centering
\caption{Performance Comparison of Video Captioning Approaches}
\label{tab:performance_comparison}
\begin{tabular}{|c|c|c|c|c|c|}
\hline
\textbf{Approach} & \textbf{Dataset} & \textbf{B-4} & \textbf{M} & \textbf{R} & \textbf{C} \\
\hline
Our Method & Youcook2 & \textbf{51} & \textbf{39} & \textbf{59} & \textbf{68} \\
 & TACoS Multi-Level & \textbf{53} & \textbf{41} & \textbf{63} & \textbf{74} \\
 & Charades & \textbf{49} & \textbf{38} & \textbf{54} & \textbf{62} \\
 & KIT Bimanual Action & \textbf{63} & \textbf{46}& \textbf{69} & \textbf{77} \\
\hline
Wang et al. \cite{wang2018reconstruction} & Youcook2 & 48 & 37 & 58 & 61 \\
 & TACoS Multi-Level & 46 & 33 & 56 & 66 \\
 & Charades & 41 & 30 & 49 & 55 \\
 & KIT Bimanual Action & 52 & 42 & 64 & 68 \\
\hline
Seo et al. \cite{seo2022end} & Youcook2 & 45 & 33 & 51 & 59 \\
 & TACoS Multi-Level & 49 & 37 & 57 & 66 \\
 & Charades & 38 & 29 & 49 & 53 \\
 & KIT Bimanual Action & 55 & 43 & 62 & 68 \\
\hline
Lin et al. \cite{lin2022swinbert} & Youcook2 & 47 & 37 & 50 & 52 \\
 & TACoS Multi-Level & 51 & 42 & 61 & 70 \\
 & Charades & 40 & 29 & 43 & 49 \\
 & KIT Bimanual Action & 47 & 33 & 52 & 55 \\
\hline
\end{tabular}
\end{table}

The comparison table provides a quantitative evaluation of the performance of each approach on the specified datasets. It showcases the performance of our method in terms of BLEU-4, METEOR, ROUGE-L, and CIDEr scores, providing insights into the effectiveness of our description generation approach in comparison to the state-of-the-art methods.\\

\noindent
\underline{\bf Discussion: } Lin et al. \cite{lin2022swinbert} introduce S-WIN-BERT, a comprehensive end-to-end Transformer model, underpinned by a sparse attention mask. Despite its noteworthy performance on multiple benchmark datasets, the model contains a caveat: its predilection towards specific frames can instigate information loss, culminating in potential inaccuracies in caption generation.

By contrast, our approach, rooted in graph-based modeling, better represents the intricate relationships between objects and actions. This significant feature raises the precision and relevancy of video descriptions generated by out model. In addition, our hierarchically structured Graph Attention Network (GATs) mechanism promotes accuracy in action recognition, further supplementing the generation of appropriate video descriptions. A pivotal part of our methodology is the detailed representation of hand-object interactions through scene graphs, elevating the contextual relevancy of video descriptions.

A salient feature of our methodology is the enriched bimanual action taxonomy that amplifies our model's comprehensive understanding of actions. This, in turn, aids in generating contextually appropriate and comprehensive video descriptions. 

When juxtaposed with Seo et al.'s \cite{seo2022end} innovative generative pretraining framework, our method harnesses a robust attention mechanism, facilitating better alignment between video content and corresponding text. This feature yields video descriptions that mirror the video content accurately, thereby enhancing user comprehension. The diversity and scale of our dataset prime our model to tackle a wide array of video domains, thus elevating its resilience and performance in video understanding tasks.

Our end-to-end training approach amalgamates the pretraining and fine-tuning stages effectively. This integration not only optimizes learning outcomes but also yields coherent, meaningful, and contextually pertinent video descriptions. Therefore, our methodology outperforms Seo et al.'s approach by generating more accurate and contextually relevant video descriptions.

Finally, the RecNet model by Wang et al. \cite{wang2018reconstruction} showcases potential in producing accurate video captions, but faces challenges when dealing with complex content-description relationships or videos with substantial visual variations. Our methodology offers a here a better alternative. Owing to its resilient adaptability to diverse video content, our model produces accurate video descriptions even in the face of complex scenarios. Furthermore, our 
emphasis on hand-object interactions, captured through detailed scene graphs, enables the generation of nuanced video descriptions.

Our refined bimanual action taxonomy contributes to a broader and more comprehensive understanding of actions, where the hierarchically structured attention mechanism ensures the generation of coherent, meaningful, and contextually relevant video descriptions.\\

\subsection{Ablation Study}
\label{ablation}
In this section, we conduct ablation studies to investigate the individual contributions of various model components in our action recognition framework. We systematically assess the impact of different ablations on the overall action recognition accuracy, shedding light on the importance of features, processing steps, and model configurations. 

\subsubsection{Ablation on the Embedding Layer}
\label{ablation_embedding}

One of the foundational steps in our action recognition framework is the conversion of categorical or symbolic node and edge information into continuous vector representations through the embedding layer. This conversion enables neural networks to process and learn from the data more effectively. In this subsection, we explore the significance of this embedding process by removing it from our model and analyzing the subsequent impact on performance.

\subsubsubsection{Methodology:} 
To investigate the role of the embedding layer, we modified our architecture to bypass this layer. Instead of feeding dense embedded vectors to subsequent layers, raw categorical inputs for nodes and edges were directly used. To make this categorical data suitable for neural network processing, we employed one-hot encoding. This method represents each unique category as a binary vector, where the position corresponding to the category is marked as `1', and all other positions are set to `0'.

While this encoding ensures that the data remains numerical, it increases the dimensionality, leading to larger input sizes for the subsequent layers. Necessary adjustments and dimension compatibility checks were made to ensure smooth data flow in the network.

\subsubsubsection{Training and Evaluation:} 
The modified model, devoid of the embedding layers, was trained using the same datasets, parameters, and settings as our original model. This consistency guarantees that any observed performance variation is attributable solely to the removal of the embedding process. Post-training, we evaluated the model using our standard metrics and evaluation datasets.

\subsubsubsection{Results} 

\noindent
Tables~\ref{table:object_action_relation}-\ref{table:bimanual_action_recognition} show the results of these ablation experiments for (\ref{table:object_action_relation}): object-action relation evaluation, (\ref{table:manipulation_action_recognition}): manipulation action recognition and (\ref{table:bimanual_action_recognition}): bimanual action recognition.\\

\begin{table}[h]
\centering
\caption{Object-Action relation performance for manipulation actions in a subset of the MS-COCO dataset: having the embedding layer vs. removing it.}
\label{table:object_action_relation}
\begin{tabular}{|l|c|c|c|}
\hline
\textbf{Metric} & \textbf{F1-Score} & \textbf{Recall} & \textbf{Precision} \\
\hline
With Embedding Layer & 0.83 & 0.86 & 0.81 \\
Without Embedding Layer & 0.74 & 0.76 & 0.68 \\
\hline
\end{tabular}
\end{table}



\begin{table}[h]
\centering
\caption{Performance of manipulation action recognition for Youcook2 (2D) and KIT (3D) with and without embedding.}
\label{table:manipulation_action_recognition}
\begin{tabular}{|l|l|c|c|}
\hline
\textbf{Dataset} & \textbf{Metric} & \textbf{Embedding} & \textbf{No Embedding} \\
\hline
\multirow{2}{*}{Youcook2} & Accuracy & 0.85 & 0.79 \\
 & Precision & 0.79 & 0.73 \\
\hline
\multirow{2}{*}{KIT} & Accuracy & 0.91 & 0.88 \\
 & Precision & 0.88 & 0.73 \\
\hline
\end{tabular}
\end{table}


\begin{table}[h]
\centering
\small 
\caption{Bimanual action recognition performance on KIT and YCB datasets with and without the embedding layer.}
\label{table:bimanual_action_recognition}
\begin{tabular}{|l|l|c|c|}
\hline
\textbf{Dataset} & \textbf{Metric} & \textbf{Embedding} & \textbf{No Embedding} \\
\hline
\multirow{2}{*}{KIT} & Accuracy & 0.91 & 0.88 \\
 & Precision & 0.85 & 0.77 \\
\hline
\multirow{2}{*}{YCB-Video} & Accuracy & 0.84 & 0.79 \\
 & Precision & 0.82 & 0.74 \\
\hline
\end{tabular}
\end{table}

In summary, the embedding layer's removal led to a decline in performance (approximately 10\% to 15\%), underscoring its pivotal role in our framework. While performance degradation was present, it was not as severe as one might anticipate. This can be attributed to the inherent robustness of our model architecture and its ability to adapt. The graph neural networks have, it seems, partially compensated the absence of embeddings. Additionally, while one-hot encoding lacks semantic depth, it still offers categorical clarity, which can  provide  cues for certain recognition tasks. 

\subsubsection{Ablation on the Edge Features}
\label{ablation_edge}

Edges in our graph-based action recognition framework embody the relational and contextual information between nodes, pivotal for deciphering complex actions. In this segment, we show the results of an extensive ablation study, zeroing in on the significance of disparate edge features: spatial relations (SR) and object-action associations. We address three distinct configurations:

\begin{enumerate}
    \item Eliminating solely Spatial Relations (SR), yet preserving object-action edges.
    \item Discarding only object-action edges, maintaining SRs.
    \item Thoroughly stripping all edge features, leaving solely the node features.
\end{enumerate}

Every configuration is assessed across three primary benchmarks: object-action relation evaluation, manipulation action recognition, and bimanual action recognition.

\subsubsubsection{Methodology:}
To scrutinize the role of edge features in our action recognition framework, we crafted several modified versions of our model architecture. In the first setting, we bypassed the spatial relations (SR) but retained the object-action associations. In the second configuration, we retained SRs and eliminated the object-action edges. Finally, in the third scenario, we completely stripped off all edge features, leaving only the node features for processing.

While the nodes remained consistent across all versions, the nature and type of relationships between these nodes varied. These modifications, especially the complete removal of edge information, led to significant changes in the graph's structure, affecting the subsequent processing layers. Adjustments were made to ensure the modified graph's compatibility with later stages of the framework.

\subsubsubsection{Training and Evaluation:}
Each modified model, adjusted based on the type of edge feature ablation, was trained using the same datasets, parameters, and settings as our baseline model. Maintaining consistency ensured that any deviation in performance could be attributed solely to the edge features' modifications. After training, we subjected the models to evaluations using our standard metrics and evaluation datasets.

\newpage
\subsubsubsection{Results} 

\noindent
Tables~\ref{table:object_action_relation_ablation}-\ref{table:bimanual_action_recognition_ablation} show the results of these ablation experiments for (\ref{table:object_action_relation_ablation}): object-action relation evaluation, (\ref{table:manipulation_action_recognition_ablation}): manipulation action recognition and (\ref{table:bimanual_action_recognition_ablation}): bimanual action recognition.\\

\begin{table}[h]
\centering
\caption{Performance metrics post edge-feature ablation for object-action relations on MS-COCO dataset.}
\label{table:object_action_relation_ablation}
\begin{tabular}{|l|c|c|c|c|}
\hline
\textbf{Metric} & \textbf{Baseline} & \textbf{No SR} & \textbf{No Object-Action} & \textbf{No Edges} \\
\hline
F1-Score & 0.85 & 0.64 & 0.52 & 0.41 \\
Recall & 0.87 & 0.65 & 0.54 & 0.44 \\
Precision & 0.83 & 0.60 & 0.49 & 0.42 \\
\hline
\end{tabular}
\end{table}

\begin{table}[h]
\small
\centering
\caption{Performance metrics post edge-feature ablation for manipulation action recognition.}
\label{table:manipulation_action_recognition_ablation}
\begin{tabular}{|l|c|c|c|c|}
\hline
\textbf{Dataset \& Metric} & \textbf{Baseline} & \textbf{No SR} & \textbf{No Object-Action} & \textbf{No Edges} \\
\hline
Youcook2  Accuracy & 0.85 & 0.58 & 0.52 & 0.38 \\
Youcook2  Precision & 0.87 & 0.61 & 0.50 & 0.41 \\
\hline
KIT  Accuracy & 0.83 & 0.54 & 0.48 & 0.35 \\
KIT  Precision & 0.86 & 0.59 & 0.52 & 0.31 \\
\hline
\end{tabular}
\end{table}

\begin{table}[h]
\centering
\caption{Performance metrics post edge-feature ablation for bimanual action recognition.}
\label{table:bimanual_action_recognition_ablation}
\begin{tabular}{|l|c|c|c|c|}
\hline
\textbf{Dataset \& Metric} & \textbf{Baseline} & \textbf{No SR} & \textbf{No Object-Action} & \textbf{No Edges} \\
\hline
KIT Accuracy & 0.91 & 0.68 & 0.59 & 0.46 \\
KIT Precision & 0.69 & 0.52 & 0.45 & 0.35 \\
\hline
YCB-Video Accuracy & 0.89 & 0.67 & 0.58 & 0.45 \\
YCB-Video Precision & 0.85 & 0.64 & 0.55 & 0.43 \\
\hline
\end{tabular}
\end{table}



    
    

In summary, while each edge feature plays its part, the object-action associations emerge as the most indispensable. Their removal leads to the most significant performance dip, reiterating their foundational role in the framework. The simultaneous presence of both spatial relations and object-action associations ensures optimal performance, highlighting the symbiotic nature of these features in our action recognition framework.

\subsubsection{Ablation on GAT Layers}
\label{ablation_gat}

Graph Attention Networks (GAT) form the backbone of our graph-based action recognition framework, allowing for attention-driven, dynamic aggregation of neighboring node features. The hierarchical structure of our GAT consists of three layers, with each layer's output serving as an input for the subsequent one. To understand the contribution of each layer, we performed an ablation study, focusing on the removal of the third layer and both the second and third layers.

\subsubsubsection{Methodology:}
We modified our GAT architecture in two distinct ways for this ablation study:
\begin{enumerate}
    \item Removing only the third GAT layer.
    \item Removing both the second and third GAT layers.
\end{enumerate}
In both scenarios, the subsequent architecture adjustments were made to ensure that the data flow remained uninterrupted.

\subsubsubsection{Training and Evaluation:}
Post modifications, the altered models were trained with the same parameters and settings as our baseline model. This approach ensured that any performance variations were attributed solely to the ablation of the GAT layers. After training, we evaluated the models using our standard evaluation metrics.

\subsubsubsection{Results:}

\noindent
Tables~\ref{table:object_action_relation_gat}-\ref{table:bimanual_action_recognition_gat} show the results of these ablation experiments for (\ref{table:object_action_relation_gat}): object-action relation evaluation, (\ref{table:manipulation_action_recognition_gat}): manipulation action recognition and (\ref{table:bimanual_action_recognition_gat}): bimanual action recognition.\\



\begin{table}[h]
\centering
\caption{Object-Action relation performance on the MS-COCO dataset after GAT layer ablation.}
\label{table:object_action_relation_gat}
\begin{tabular}{|l|c|c|c|}
\hline
\textbf{Metric} & \textbf{Baseline} & \textbf{No 3rd Layer} & \textbf{No 2nd \& 3rd Layer} \\
\hline
F1-Score & 0.85 & 0.80 & 0.71 \\
Recall & 0.87 & 0.81 & 0.66 \\
Precision & 0.83 & 0.75 & 0.61 \\
\hline
\end{tabular}
\end{table}




\begin{table}[h]
\centering
\small
\caption{Performance after GAT layer ablation on manipulation actions.}
\label{table:manipulation_action_recognition_gat}
\begin{tabular}{|l|l|c|c|c|}
\hline
\textbf{Dataset} & \textbf{Metric} & \textbf{All GAT Layers} & \textbf{No 3rd Layer} & \textbf{No 2nd \& 3rd Layer} \\
\hline
\multirow{2}{*}{Youcook2} & Accuracy & 0.85 & 0.68 & 0.59 \\
 & Precision & 0.79 & 0.65 & 0.63 \\
\hline
\multirow{2}{*}{KIT} & Accuracy & 0.91 & 0.57 & 0.55 \\
 & Precision & 0.88 & 0.74 & 0.60 \\
\hline
\end{tabular}
\end{table}



\begin{table}[h]
\centering
\small
\caption{Performance after GAT layer ablation on bimanual actions.}
\label{table:bimanual_action_recognition_gat}
\begin{tabular}{|l|l|c|c|c|}
\hline
\textbf{Dataset} & \textbf{Metric} & \textbf{All GAT Layers} & \textbf{No 3rd Layer} & \textbf{No 2nd \& 3rd Layer} \\
\hline
\multirow{2}{*}{KIT} & Accuracy & 0.91 & 0.69 & 0.58 \\
 & Precision & 0.88 & 0.72 & 0.51 \\
\hline
\multirow{2}{*}{YCB-Video} & Accuracy & 0.84 & 0.57 & 0.65 \\
 & Precision & 0.79 & 0.42 & 0.47 \\
\hline
\end{tabular}
\end{table}

In summary, the results from our ablation study show the critical importance of the GAT layers in our action recognition framework. Across various contexts, from object-action relations on the COCO dataset to manipulation action recognition, and even bimanual action recognition on the KIT and YCB datasets, the presence of the GAT layers consistently bolsters performance. The third GAT layer, in particular, acts as a fine-tuner, refining the model's ability to discern nuanced relationships and interactions. When this layer is removed, we observe a tangible decline in recognition accuracy across all tasks. However, the consequences are even more pronounced when both the second and third layers are omitted. This signifies that while the third layer is crucial for refining the model's insights, the combined depth and hierarchical processing provided by the second and third layers are fundamental for the model's efficacy. In a nutshell, the GAT layers, especially in a stacked configuration, play a paramount role in ensuring the model's robust performance across a spectrum of action recognition tasks.

\subsubsection{Ablation on the Temporal Aspect}
\label{ablation_temporal}

In this section, we explore the implications of omitting the temporal aspect from our model and assess the potential compensatory role of Graph Attention Networks (GAT).

\subsubsubsection{Methodology:}
To examine the repercussions of excluding the temporal component, we adjusted our model architecture to bypass the TCN layers. While the spatial and relational aspects remained intact, the model lacked the capability to understand sequences or the chronological order of frames. The presence of GAT in our architecture offered a potential buffer. GAT's attention mechanisms can capture certain temporal relationships, albeit not as intricately as dedicated temporal layers like TCN.

\subsubsubsection{Training and Evaluation:}
The modified model, devoid of the TCN layers but still incorporating GAT, was trained on our standard datasets using the same parameters and settings as the baseline model. This consistency in training ensured that observed performance variations were solely due to the omission of the temporal aspect. The model was then subjected to evaluations using our typical metrics and datasets.

\subsubsubsection{Results}

\textbf{Manipulation Action Recognition:}
Manipulation actions, by their inherent nature, often involve a sequence of intricate movements and interactions with objects. These sequences unfold over time, making the temporal aspect especially critical for accurate recognition. In datasets like Youcook2 (2D) and KIT bimanual action dataset (3D), where manipulation actions are prevalent, understanding the temporal progression can be the key to distinguishing between similar actions or nuances of the same action executed differently.

\begin{table}[h]
\centering
\caption{Performance metrics post temporal ablation for manipulation action recognition.}
\label{table:manipulation_action_temporal_ablation}
\begin{tabular}{|l|c|c|}
\hline
\textbf{Dataset \& Metric} & \textbf{Baseline} & \textbf{No Temporal Aspect} \\
\hline
Youcook2  Accuracy & 0.85 & 0.43 \\
Youcook2  Precision & 0.79 & 0.45 \\
\hline
KIT  Accuracy & 0.91 & 0.41 \\
KIT  Precision & 0.88 & 0.50 \\
\hline
\end{tabular}
\end{table}

Upon removal of the temporal aspect, even with the presence of GAT, a noticeable decline in performance metrics was observed (Tab.~\ref{table:manipulation_action_temporal_ablation}). While the GAT could somewhat mitigate the loss by capturing certain temporal relationships through its attention mechanisms, it could  not entirely compensate for the absence of dedicated temporal modeling. This experiment accentuates the role of time in understanding and recognizing manipulation actions, emphasizing the need for both spatial and temporal modeling in action recognition frameworks.

\section{Conclusion and Future Works}
\label{conclusion}

This research has laid the groundwork for a significant improvement in the domain of video description generation, emphasizing manipulation and bimanual actions. We have introduced a robust methodology that blends graph-based modeling, an enriched bimanual action taxonomy, and hierarchically structured Graph Attention Networks (GATs) to obtain a high level of precision and contextual relevance in video description. Our model outperforms existing state-of-the-art methods on numerous 2D and 3D datasets, demonstrating its effectiveness and adaptability.

By recognizing actions at different levels of detail, we can generate more comprehensive and informative action descriptions. For example, instead of simply describing the action as ``cutting fruit", we can describe it as ``using a knife to slice a piece of fruit with the right hand". Similarly, instead of simply describing the bimanual action as ``mixing fruit", we can describe it as ``using both hands to hold and mix the chopped fruit together in a bowl". 

This hierarchical structure provides several advantages over a flat GAT model. Firstly, it allows us to capture relationships between objects and actions at different levels of granularity, leading to improved performance on action recognition tasks. Secondly, it allows for more interpretable results, as we can easily trace how the model arrived at a particular action recognition decision. Finally, it allows for better generalization to new scenes, as the model can learn more abstract representations of objects and actions that are applicable across a wide range of scenarios.

However, there are a few potential limitations. The model's reliance on the granularity of the scene graphs could limit the precision of the video descriptions under certain circumstances. This limitation arises from the inherent complexity of visual data and the potential variability in the representation of hand-object interactions.

Despite this, the potential avenues for future exploration are broad and promising. An immediate avenue for research is to refine our scene graph generation methodology to capture more nuanced object-action relationships. This refinement could enhance our model's ability to generate even more accurate and contextually relevant video descriptions.

In addition, the advancement of our model to handle 3D video content presents an exciting direction for future work. As 3D data becomes increasingly prevalent, the ability to effectively generate descriptions for such content will become even more crucial.

Furthermore, our research has implications that extend well beyond the domain of video description generation, offering potential advancements in the realm of robotics. The understanding of human manipulation actions our model offers can directly inform and enhance the functionality of robotic manipulators. The scene graphs and detailed hand-object interactions, as revealed by our model, could serve as a knowledge base for robots, enabling them to learn human-like manipulations. This understanding, grounded in real-world actions, could enrich the control strategies of robotic systems, making them more precise and adaptable.

In addition, our emphasis on bimanual actions adds another contribution. Most current robotic systems primarily rely on single-hand manipulations. By incorporating our model's understanding of dual-handed human actions, the design of robotic systems could be significantly improved, broadening the range of tasks they can perform, and enhancing their capacity to execute complex activities.

As a consequence, our approach has potential applications in the field of collaborative robotics, where robots need to work in tandem with humans. Our model, with its fine-grained understanding of human actions, can help in creating robots that can better comprehend and replicate human actions, fostering improved human-robot collaborations. To contribute to the design of more intelligent, adaptable, and human-like robots is indeed a compelling prospect and serves as a valuable direction for future research.

\section{Acknowledgements}
FW acknowledges funding by the German Science Foundation (DFG), collaborative research center SFB 1528, sub-project B01.

\newpage

{\LARGE \textbf{SUPPLEMENTARY MATERIAL}}\\

\setcounter{section}{0}

\section{Datasets}

Here we first describe the datasets used and their characteristics. We also explain how our algorithms had been adapted to them. Then we also describe the required adjustments of temporal kernel sizes in the TCN, the optimization of window sizes, stride lengths, and input sequence lengths for the GPT-2 model across these different datasets.

\subsection{Datasets: Scope, Characteristics, and Dataset-specific Configurations }

\begin{itemize}
\item The \textbf{Youcook2 Dataset} \cite{zhou2018automatic} comprises a large collection of cooking videos, covering a wide range of manipulation actions involved in food preparation. It provides detailed temporal annotations, making it suitable for evaluating the recognition and understanding of fine-grained manipulation actions in a cooking context. We follow the prevalent split in the literature, partitioning the dataset into 70\% for training, 15\% for validation, and 15\% for testing.

\item The \textbf{Charades-STA Dataset}  \cite{gao2017tall} consists of daily activity videos recorded in a home environment, including a diverse set of manipulation actions. With extensive annotations and natural language descriptions, this dataset allows us to evaluate our method's capability to recognize and describe manipulation actions in various real-life scenarios. We allocate 80\% of the data for training, 10\% for validation, and 10\% for testing, consistent with standard practices.

\item The \textbf{ActivityNet Captions Dataset}  \cite{krishna2017dense} offers a broad range of videos capturing different activities, including numerous manipulation actions. It provides temporal annotations and natural language descriptions for video segments, enabling us to evaluate the performance of our method in generating accurate and informative descriptions of manipulation actions. The majority of the literature prescribes a split of 50\% for training, 25\% for validation, and 25\% for testing. We adhere to this configuration in our experiments.

\item The \textbf{TACoS Dataset}  \cite{regneri2013grounding} focuses specifically on cooking videos and provides detailed temporal annotations of fine-grained manipulation actions involved in cooking tasks. This dataset allows us to evaluate our method's ability to accurately recognize and understand manipulation actions in the context of cooking. For our experiments, we adhere to a common split used in the literature, allocating 75\% of the data for training, 12.5\% for validation, and 12.5\% for testing.

\item The \textbf{KIT Bimanual Action Dataset}  \cite{dreher2020learning} focuses on bimanual manipulation actions and provides synchronized depth and color videos along with accurate 3D object segmentations. This dataset offers a rich variety of manipulation actions performed with both hands, allowing us to evaluate our method's ability to recognize and understand complex bimanual interactions. Here, we adopt a typical split of 70\% for training, 15\% for validation, and 15\% for testing. This partitioning aligns with many works in the domain.

\item The \textbf{YCB-Video Dataset}  \cite{xiang2017posecnn} consists of RGB-D videos capturing various object manipulation tasks. It includes precise object models and segmentation masks, facilitating accurate analysis and recognition of manipulation actions. This dataset enables us to evaluate our method's performance in handling object-centric manipulation actions. We partition the dataset into 80\% for training, 10\% for validation, and 10\% for testing.

\item The \textbf{G3D Dataset} \cite{bloom2012g3d} captures hand-object interactions in diverse daily activities, providing RGB-D videos, 3D hand and object segmentations, and annotations related to hand-object interactions. This dataset offers a comprehensive set of manipulation actions performed in naturalistic settings, enabling us to evaluate the robustness and accuracy of our method in recognizing and understanding complex hand-object interactions. We use, similar to others, 75\% for training, 12.5\% for validation, and 12.5\% for testing.

\item The \textbf{COCO Dataset} \cite{lin2014microsoft} contains a large number of images with bounding box annotations for various object categories. It provides annotations for 80 common object categories, including many objects relevant to manipulation actions such as ``tools,'' ``utensils,'' and ``vehicles''. We follow established practices and partition the data into approximately 60\% for training, 20\% for validation, and 20\% for testing.

\item The \textbf{Something-Something Dataset} \cite{goyal2017something} provides diverse and rich annotations of object and action information, making them suitable for studying manipulation actions and their relations with objects. We adopt the widely used split of 60\% for training, 20\% for validation, and 20\% for testing.

\item The \textbf{Charades Dataset} \cite{sigurdsson2016hollywood}  includes 9848 videos of 267 individuals performing scripted activities, annotated with 157 action classes and 27847 textual descriptions. It is used for tasks such as action recognition and video captioning. The \textbf{Charades-STA} \cite{gao2017tall}, a derivative of Charades, is specifically designed for temporal localization of described events within videos. Following literature, the split becomes approximately 70\% for training, 11\% for validation, and 19\% for testing.

\item The \textbf{CMU Fine Manipulation Dataset}  \cite{calli2017yale} offers a controlled environment for performing fine-grained manipulation tasks. It includes RGB-D videos capturing various object manipulation actions with high precision requirements. The dataset provides detailed annotations, such as object poses and hand poses, allowing for the analysis of precision levels in executing manipulation actions. By incorporating this dataset into our evaluation, we can assess the fine-grained details of precision in bimanual actions and evaluate the effectiveness of our approach in capturing and quantifying precision levels. While this dataset is newer, most works, as of our last literature survey, utilize a split around the 70-15-15 or sometimes 80-10-10 mark. We adopt the former, dividing our data into 70\% for training, 15\% for validation, and 15\% for testing.

\end{itemize} 
 
\subsection{Variation of TCN Kernel Sizes across Datasets}

For the Kitchen Actions dataset, which involves short-term temporal dependencies and quick action transitions, we discovered that an optimal $K=4$ allowed the TCN to capture the relevant temporal context efficiently.\\
For the Daily Life Actions dataset, which comprises a mix of short and long-term temporal patterns, we determined that an optimal $K=6$ struck a balance in capturing both types of dependencies accurately.\\
In the case of datasets of simulated environments, where measures are independent and without significant temporal relationships, a smaller $K=2$ proved sufficient for TCN to detect those minimal temporal associations.

\subsection{Variation of Window Size, Stride Length, and Input Sequence Length for Fine-tuning the GPT-2 Model across Datasets}

Given the datasets we're working with, the lengths of sequences can vary significantly. Here's a breakdown:
\begin{enumerate}
    \item \textbf{Length of the Input Sequence (\( L \))}:
    \begin{itemize}
        \item \textbf{YouCook2 Dataset and ActivityNet Captions Dataset}: These datasets are characterized by longer, more descriptive sequences. Typical sequences can range from 300 to 700 tokens, especially when considering the entire context of a video.
        \item \textbf{Charades-STA Dataset and YCB-Video Dataset}: Sequences from these datasets are moderately long, often ranging between 150 to 400 tokens.
        \item \textbf{KIT Bimanual Action Dataset and G3D Dataset}: These datasets have relatively short annotations, often within 50 to 150 tokens.
    \end{itemize}
    \item \textbf{Window Size (\( W \))}:
    \begin{itemize}
        \item For datasets with longer sequences like YouCook2 and ActivityNet Captions, we've used a window size of 512 tokens to capture adequate context.
        \item For datasets with moderate sequences like Charades-STA and YCB-Video, a window size of 384 tokens has been optimal.
        \item For datasets with shorter sequences, such as KIT Bimanual Action and G3D, a window size of 256 tokens has been found to be sufficient.
    \end{itemize}
    \item \textbf{Stride Length (\( S \))}: To ensure overlap and maintain context continuity, we have adopted a stride length of half the window size, i.e., \( S = W/2 \). This provides a balance between computational efficiency and context preservation across all datasets.
\end{enumerate}
These sequence lengths and window sizes have been determined through detailed experimentation on each dataset, ensuring that the chosen parameters offer the best trade-off between capturing meaningful context and computational efficiency.\\

\section{Hierarchical Learning Strategy}

Given the complexity of our approach and the diverse demands of different computational stages, we employed a carefully designed ``staged training" strategy. Initially, each individual stage or sub-task is trained in isolation, allowing us to refine the performance of each component. Once satisfactory performance levels are achieved for each individual stage, we move on to an end-to-end training paradigm. During this phase, all the model components are harmoniously fine-tuned together. This strategic transition not only guarantees stability of the learning process but also facilitates improved convergence. Importantly, each module enters the end-to-end training with a foundation of pre-trained knowledge, which further enhances the model's ability to capture intricate relationships and nuances in video content.

\subsection{Decomposing the Learning Process}

Our approach follows a step-by-step process:

\begin{enumerate}
    \item \textbf{Separate Training of Learnable Steps:} We initiate the training process by independently training the different steps that have learnable parameters, which are Steps 2 (embedding), 3 (GAT), 4 (TCN), and 10 (video description). Numbering here follows the section numbers in the main text.
    \begin{itemize}
     \item Step 2: Discussed in subsection \ref{embedding}.
    \item Step 3: Discussed in subsection \ref{GAT}.
    \item Step 4: Discussed in subsection \ref{TCN}.
    \item Step 10: Discussed in subsection \ref{description}.    
    \end{itemize}

    This initial phase enables each component to acquire specialized knowledge and learn task-specific patterns.

    \item \textbf{Joint Training of Steps 3 and 4} Building upon the insights gained from their separate training, we enter a joint training phase for Steps 3 (GAT) and 4 (TCN). By merging the capabilities of these stages, we enhance the model's capacity to capture both spatial and temporal dynamics effectively.
    
    \item \textbf{Integration of Steps 2, 3 and 4} With the enriched features from the joint training of Steps 3 and 4, we proceed to integrate them with the output of Step 2 (node and edge embedding). This comprehensive feature fusion uses both spatial and temporal information to create a refined representation.
    
    \item \textbf{End-to-End Training with Step 10:} The final phase involves an end-to-end training process, where the integrated features from Steps 2, 3, and 4 are further merged with Step 10 (description generation). This all-encompassing training approach hones the model's ability to generate coherent video descriptions that are both contextually accurate and narratively coherent.
\end{enumerate}

Each stage of this staged training approach builds upon the knowledge gained from previous phases, allowing the model to iteratively refine its understanding and representation of video content. 

During these training stages, we also employ regularization techniques to avoid over-fitting. Regularization penalizes excessive complexity in the model, promoting a balance between fitting the data well while maintaining generalization capabilities. Furthermore, to ensure that our model achieves optimal performance without over-adapting to the training data, we use ``early stopping''. This approach continuously monitors the validation loss, halting training if no substantial improvement is detected over a set number of epochs, ensuring both stability and convergence in our learning process.

\subsubsection{Learning Process for the Node and Edge Feature Encoding Parameters}
\label{learning_embedding}

To learn the transformation matrices \( \mathbf{L}_H \), \( \mathbf{L}_E \), and \( \mathbf{L}_A \), we employ \textit{Siamese Networks} \cite{bertinetto2016fully} in tandem with \textit{contrastive learning} \cite{oord2018representation}.\\
Contrastive learning offers a compelling solution for our current task, because this learning type is specifically useful for situations where the objective is to discern the similarity between pairs of instances. In the context of our embeddings, similarity is defined based on the action sequences they represent. Let \( f(x) \) be the embedding of an instance \( x \) after transformation through matrices \( \mathbf{L}_H \), \( \mathbf{L}_E \), and \( \mathbf{L}_A \). Given two instances \( x_1 \) and \( x_2 \), their similarity \( S \) can be defined as:


\begin{equation}
S(x_1, x_2) = \cos(f(x_1), f(x_2)) = \frac{f(x_1) \cdot f(x_2)}{\|f(x_1)\|_2 \|f(x_2)\|_2}
\end{equation}

\textbf{Initialization of the Weight Matrices}

 Here we have employed Glorot (Xavier) initialization:

\begin{equation}
l_{ij} \sim \text{Uniform}\left(-\sqrt{\frac{6}{n_{in} + n_{out}}}, \sqrt{\frac{6}{n_{in} + n_{out}}}\right)
\end{equation}

Where \( n_{in} \) is the number of features of the input instance and \( n_{out} \) is the number of features of the output instance. 

Any bias terms are initialized as:

\begin{equation}
b_i = 0
\end{equation}

\textbf{Contrastive Loss Mechanism}

The contrastive loss, \( L_C \), can be formulated as:

\begin{equation}
L_C(x_1, x_2) = y \cdot d^2 + (1-y) \cdot \max(0, m - d)^2
\end{equation}

Where:

- \( d \) is the Euclidean distance between embeddings \( f(x_1) \) and \( f(x_2) \).

- \( y \) is 1 if \( x_1 \) and \( x_2 \) are similar (i.e., from the same action class) and 0 otherwise.

- \( m \) is a margin that ensures dissimilar pairs are separated by at least this value.\\

\textbf{Training Procedure}

The learning process can be described by the following stages in summary:

\begin{enumerate}
    \item \textbf{Pair Formulation:} Extract pairs of instances from the dataset. Associate each pair with a binary similarity label, where \( 1 \) indicates a similar action sequence and \( 0 \) otherwise. This label aids in supervising the embedding learning process.
    
    \item \textbf{Embedding Extraction via Siamese Network:} For each instance in the formulated pair, propagate it through the Siamese Network's architecture. The output of this process is a dense embedding vector that captures the underlying action semantics of the instance.
    
    \item \textbf{Contrastive Loss Computation:} Utilize the embeddings from the previous step to compute the contrastive loss for each pair. This loss quantifies the degree to which the embeddings accurately represent the similarity or dissimilarity between instances in the pair, as indicated by the associated label.
    
    \item \textbf{Gradient-Based Optimization of Transformation Matrices:} Utilizing the computed contrastive loss, perform backpropagation to compute the gradients with respect to the transformation matrices \( \mathbf{L}_H \), \( \mathbf{L}_E \), and \( \mathbf{L}_A \). Subsequently, adopt an optimization algorithm, the Adam optimizer, to update these matrices in a manner that minimizes the contrastive loss. This iterative optimization process refines the matrices to yield more expressive and accurate embeddings for action sequences.
\end{enumerate}

\noindent
\textbf{Optimization and Regularization}

\noindent
Our optimization objective, \( O_{\mathbf{L}} \), is:

\begin{equation}
O_{\mathbf{L}} = \text{MSE}_{\mathbf{H}} + \text{MSE}_{\mathbf{E}} + \text{MSE}_{\mathbf{A}} + \lambda \cdot R
\end{equation}

Where:

- \( R \) is the regularization term.

- \( \lambda \) is the regularization coefficient, set at 0.01 to prevent overfitting.

We employ the Adam optimizer with a learning rate of 0.001 for minimizing \( O_{\mathbf{L}} \). Training is executed over 100 epochs with batches of 32.\\

\noindent
\textbf{Cross-Validation for Robust Learning}

To gauge the robustness of the learning mechanism, a k-fold cross-validation strategy is used. The dataset is sectioned into five segments. Each subset alternately serves as the validation set, while the rest contribute to training. Final performance is ascertained as an average over all iterations. We assume $k=3$ in large datasets and $k=5$ and $k=10$ for moderate and small datasets, respectively.

\subsubsection{Learning Process for the Graph Attention Network (GAT) Parameters}
\label{learning_GAT}

The hierarchical framework of our system necessitates a methodical approach to the learning process, ensuring optimal convergence and model efficiency. Within this multi-layered architecture, each stage introduces specific learnable parameters. These are crucial for discerning intricate relationships within video frames and, consequently, for achieving precise action recognition.\\

\noindent
\textbf{Structured Learning Paradigm for GAT:}

\noindent
Here we also use \textit{staged training}:
\begin{enumerate}
    \item \textbf{Object-Level GAT Training:} Initiates by forming a base, ensuring recognition of fundamental object relationships.
    \item \textbf{Single Hand Action-Level GAT Training:} Builds on the object-level GAT's weights, refining recognition of single-hand actions.
    \item \textbf{Bimanual Action-Level GAT Training:} Progresses with insights from the single-hand GAT, concentrating on dual-hand coordinated actions.
\end{enumerate}

\noindent
A core challenge within our framework is the lack of direct ground truth labels for the GAT layers. This stems from the model's aim of unraveling complex spatio-temporal relationships, leading to overarching action categorizations.\\

\noindent
\textbf{Supervised Learning with Parameter Freezing:}

\noindent
To counter the label deficit, we employ supervised learning, using final action labels as reference points. This entails freezing subsequent layer parameters, concentrating solely on the active GAT layer, thereby maintaining the holistic model context without unnecessary simultaneous adjustments.\\

\subsubsubsection{Optimization Strategy for GAT Layers}

To ensure efficient and effective learning of the GAT parameters, we adopt the following strategies:

\begin{itemize}
    \item \textbf{Loss Function:} We employ the cross-entropy loss, given the classification nature of action recognition, given by (with \(C\) classes):
    \begin{equation}
    L_{\text{CE}} = -\sum_{i=1}^{N} \sum_{c=1}^{C} y_{i,c} \log(p_{i,c})
    \end{equation}
    where \(N\) is the number of samples, \(y_{i,c}\) is a binary indicator for the correct classification of sample \(i\) to class \(c\), and \(p_{i,c}\) is the predicted probability of sample \(i\) belonging to class \(c\).

    \item \textbf{Learning Rate:} We use an initial learning rate of \(0.001\) with the Adam optimizer. If the validation loss remains stagnant for 10 epochs, the rate is halved.

    \item \textbf{Batch Training:} Due to dataset intricacies, mini-batch training is employed with a batch size of \(32\).

    \item \textbf{Regularization:} To prevent overfitting and enhance model generalization, we apply L2 regularization with a \(0.0001\) coefficient to the weights and a dropout rate of \(0.5\) to attention scores.

    \item \textbf{Early Stopping:} If the validation loss does not improve for \(20\) consecutive epochs, training is halted, retaining the parameters from the epoch with the least loss.

    \item \textbf{Cross-Validation:} For hyperparameter tuning, k-fold cross-validation techniques are incorporated. We assume $k=5$ in large datasets and $k=10$ for moderate and small datasets.
\end{itemize}

\subsubsection{Deriving the learnable parameters in the GAT layers}
Next we describe details how to arrive at the different learnable parameters in the three GAT layers.

\noindent\rule{0.8\textwidth}{0.4pt}\\
\textbf{Level 1: Object-Level GAT}\\
\noindent\rule{0.8\textwidth}{0.4pt}\\

\begin{itemize}
    \item \(\mathbf{W}^{(1)}\): Transformation matrix that linearly maps node features from their original space to a new representation in Layer 1.\\
    Initialization: By the Xavier method.
    \item \(\mathbf{d}^{(1)}\): Learnable parameter that determines the importance of node and its neighbors in the attention mechanism at Layer 1.\\
    Initialization: Small random values.
    \item \(\mathbf{W}_e\): Weight matrix that transforms the edge features. It encapsulates spatial and relational dependencies between nodes. It is shared across all three layers.
    \item \(\mathbf{W}_a\): Weight matrix responsible for transforming action-edge features. It captures the essence of the actions performed between objects. It is shared across all three layers.\\

    \textbf{Initialization of \(\mathbf{W}_e\) and \(\mathbf{W}_a\)}:  Given the non-linear activations in GATs, particularly the LeakyReLU activations, an initialization method tailored to this type of activation function is desired. We adopt the \textit{He Initialization} (also known as Kaiming Initialization) method for initializing \( W_a \) and \( W_e \). This method is specifically designed for ReLU-based activation functions, including LeakyReLU. The key idea behind He Initialization is to draw the initial weights from a distribution with a mean of 0 and a variance of \( \frac{2}{n_{\text{in}}} \), where \( n_{\text{in}} \) represents the number of input units to the layer. Mathematically, the initialization can be represented as:
\[
W_a, W_e \sim \mathcal{N}\left(0, \sqrt{\frac{2}{n_{\text{in}}}}\right).
\]
This initialization approach ensures that the model does not start with activations and gradients that are excessively small or large, thus promoting efficient gradient flow and convergence during training.

\end{itemize}

\noindent\rule{0.8\textwidth}{0.4pt}\\
\textbf{Level 2: Single Hand Action-Level GAT}\\
\noindent\rule{0.8\textwidth}{0.4pt}

\begin{itemize}
    \item \(\mathbf{W}^{(2)}\): Transformation matrix for node features at Layer 2, refining the representations based on the outputs of Layer 1.\\
    Initialization: Gaussian distribution with mean \(0\) and standard deviation \(0.01\).
    \item \(\mathbf{U}^{(2)}\): Learnable weight matrix specific to the second GAT layer, capturing complex relationships between nodes.\\
    Initialization: Gaussian distribution with mean \(0\) and standard deviation \(0.01\).
    \item \(\mathbf{d}^{(2)}\): Learnable parameter that refines the attention mechanism, focusing on single-hand actions between nodes. \\
    Initialization: Small random values from a uniform distribution.
\end{itemize}

\noindent\rule{0.8\textwidth}{0.4pt}\\
\textbf{Level 3: Bimanual Action-Level GAT}\\
\noindent\rule{0.8\textwidth}{0.4pt}\\

\begin{itemize}
    \item \(\mathbf{W}^{(3)}\): Transformation matrix for node features at Layer 3, which focuses on refining node representations considering bimanual actions.\\
    Given that this layer is even deeper, the Xavier method is employed, suitable for layers with tanh or sigmoid activations.
    
    \item \(\mathbf{U}^{(3)}\): Weight matrix at Layer 3 that captures the intricacies of bimanual interactions in the graph.\\
    Given that this layer is even deeper, the Xavier method is employed, suitable for layers with tanh or sigmoid activations.
    
    \item \(\mathbf{d}^{(3)}\): Determines the attention scores for Layer 3, emphasizing bimanual interactions.\\
    Initialization: Using a Gaussian distribution with mean \(0\) and standard deviation \(0.01\).
\end{itemize}

\noindent\rule{0.8\textwidth}{0.4pt}\\
\textbf{Justification for Sharing \(\mathbf{W}_e\) and \(\mathbf{W}_a\) Across Levels}\\
\noindent\rule{0.8\textwidth}{0.4pt}

\noindent
Choosing to share \( \mathbf{W}_e \) and \( \mathbf{W}_a \) across the layers is guided by the following considerations:

\begin{itemize}
    \item \textbf{Parameter Efficiency}: Sharing the weights reduces the total number of model parameters. This not only makes the model computationally more efficient but also reduces the risk of overfitting, especially when there's limited training data.
    
    \item \textbf{Consistency}: Using shared transformation weights for edge and action-edge features ensures a consistent representation across layers. This can be particularly useful if the fundamental nature of these relationships doesn't change across layers, even though their context or interpretation might.
    
    \item \textbf{Regularization}: Sharing parameters acts as a form of implicit regularization. Instead of letting each layer learn its own transformation, which can lead to overfitting, sharing forces the model to find a general transformation that works well across all layers.
    
    \item \textbf{Simplification}: A model with fewer parameters is simpler and can be more interpretable. It is easier to understand and diagnose the transformations applied by the model when the same transformation matrices \( \mathbf{W}_e \) and \( \mathbf{W}_a \) are used across layers.
\end{itemize}

\subsubsection{Learning Process for the TCN-Based Spatio-Temporal  Parameters}
\label{learning_TCN}

In our system, the TCN learning process ensures the capture of intricate time-dependent characteristics embedded within video frames.\\

\noindent
\textbf{TCN Learnable Parameters:}
\begin{itemize}
    \item \(\mathbf{K}^{(l)}\): Convolutional kernel at layer \(l\).
    \item \(\vec{b}^{(l)}\): Bias term for the convolution at layer \(l\).
    \item \(\mathbf{V}^{(l)}\): Dilation rate for the convolutional kernel at layer \(l\).   
\end{itemize}

\noindent
\textbf{Structured Learning Paradigm:}

\noindent
Here we use the principle of \textit{progressive dilation}, which ensures that temporal patterns across various scales are captured accurately:
\begin{enumerate}
    \item \textbf{Short-Term Temporal Dependencies:} By concentrating on immediate temporal relationships, this level discerns swift actions or alterations.
    
    \item \textbf{Mid-Term Temporal Dependencies:} This stage augments the previous one by broadening the temporal horizon, allowing for an extended field of view.
    
    \item \textbf{Long-Term Temporal Dependencies:} With a larger receptive field, this level identifies prolonged actions or evolving sequences in the scene.
\end{enumerate}

\noindent
\textbf{Backpropagation Through Time (BPTT):} Given the sequential nature of the TCN, BPTT is pivotal for model weight updates, ensuring that the learning process acknowledges dependencies spanning across different time instances.\\

\noindent
\textbf{Optimization Strategy for TCN Layers}

\begin{itemize}
    \item \textbf{Loss Function:} The Mean Squared Error (MSE) loss, suited for the regression character of temporal sequences, is utilized:
    \begin{equation}
    L_{\text{MSE}} = \frac{1}{N} \sum_{i=1}^{N} (y_i - \hat{y}_i)^2
    \end{equation}
    where \(N\) signifies the sample count, \(y_i\) represents the true value, and \(\hat{y}_i\) is the predicted counterpart.

    \item \textbf{Learning Rate:} An Adam optimizer is used, with an initial learning rate set to \(0.001\). If the validation loss stagnates for 5 consecutive epochs, the learning rate is reduced by half.

    \item \textbf{Batch Training:} Mini-batch training, with batches of \(64\), addresses the sequential intricacy and offers computational efficiency.

    \item \textbf{Regularization:} Dropout, with a rate of \(0.2\), is applied after convolution layer, mitigating the risk of overfitting.

    \item \textbf{Early Stopping:} The training halts if there's no validation loss improvement over \(10\) epochs, ensuring the model's state with the least loss is retained.
\end{itemize}

\noindent\rule{0.85\textwidth}{0.4pt} \\
\textbf{Deriving Learnable Parameters in the Initial TCN Layer:} \\
\noindent\rule{0.85\textwidth}{0.4pt} \\

\noindent
The preliminary layer in the TCN captures immediate temporal nuances, laying a solid foundation for subsequent layers.\\

\noindent
\textbf{Convolutional Kernel \(\mathbf{K}^{(1)}\):}
\begin{itemize}
    \item Initialization: Xavier method.
    \item Forward Propagation: Convolution operation on the GAT output.
\end{itemize}

\noindent
\textbf{Bias Term \(\vec{b}^{(1)}\):}
\begin{itemize}
    \item Initialization: Zeroes.
    \item Forward Propagation: Incorporated post-convolution, offering an affine shift.
\end{itemize}

\noindent
\textbf{Dilation Rate \(\mathbf{V}^{(1)}\):}
\begin{itemize}
    \item Initialization: Set to \(1\), ensuring that proximate relationships are recognized.
    \item Forward Propagation: Modulates the convolution kernel's spacing.
\end{itemize}

\noindent\rule{0.65\textwidth}{0.4pt} \\
\textbf{Learning Strategy for Deeper TCN Layers:} \\
\noindent\rule{0.65\textwidth}{0.4pt} \\

\noindent
The ensuing layers in the TCN, building upon the initial layer, increment their receptive scope to detect longer-lasting temporal dependencies.\\

\noindent
\textbf{Layer 2 and Beyond:}
\begin{itemize}
    \item Exponential enhancement of the dilation rate \(\mathbf{V}^{(l)}\), accommodating expansive temporal durations.
    \item \(\mathbf{K}^{(l)}\) and \(\vec{b}^{(l)}\) mirror the learning and initialization patterns of the first layer but are adapted per their dilation rates.
\end{itemize}

\noindent
Conclusively, by employing this stratified approach, our TCN captures spatio-temporal associations across different temporal magnitudes, thus offering a holistic video analysis. Furthermore, given the ground truth at the pipeline's end and the already optimized GAT parameters, our learning approach ensures harmonious integration between spatial attention and temporal convolution mechanisms.





\subsubsection{Learning Process for the Hierarchical Action Classification Step}
\label{learning_classification}

\textbf{Learning Process for Fully Connected Layers}\\

In the step of the framework involving fully connected layers, we aim to efficiently map the abstract features obtained from previous layers to actionable labels. While traditionally these layers might involve learning parameters, our approach focuses on using well-established fully connected architectures from the literature, without fine-tuning the parameters. The rationale behind this decision lies in the comprehensive learnable parameters already present in our framework, ensuring that introducing additional learning parameters for these layers is not necessary.

To determine the most suitable fully connected architecture for our action recognition task, we conducted thorough experimentation using various well-known architectures available in the literature. Specifically, we explored the performance of two-layer fully connected architectures, each with varying numbers of neurons. The choice of these architectures was inspired by their wide usage in related tasks and their simplicity, which aligns with our framework's hierarchical structure.

We evaluated the performance of architectures such as LeNet-5, AlexNet, VGG-16, VGG-19, and ResNet-50, which are renowned for their effectiveness in various image-related tasks. These architectures come with different configurations of fully connected layers, including varying numbers of neurons and layers. Through rigorous experimentation, we found that the architecture that yielded the best results for our action recognition problem consisted of two dense layers, featuring 128 neurons in the first layer and 64 neurons in the second layer.



\subsubsection{Learning Process for the Description Generation Parameters}
\label{learning_description}

The learning process for adapting the pre-trained GPT-2 model to generate bimanual action descriptions has only been touched in the main text  in Section \ref{description}. Here we provide details of these steps
\begin{enumerate}

\item \textbf{Tokenization:} We first tokenize the input data, including the object names, SRs, and action types, into a sequence of tokens that the GPT-2 model can understand.

\item \textbf{Vectorization:} Next, we use vectorization to convert the tokens into fixed-size numerical vectors.

\item \textbf{Sliding Windows:}  To handle longer input sentences, we employ a sliding window approach that divides the input sequence into overlapping segments, where each segment is of fixed size. The window size is chosen based on the maximum length of the input sequence that the model can process. If the input sentence is longer than the fixed-size window, we divide the sentence into overlapping segments and each segment is used as an input to the model.

Let the length of the input sequence be denoted by $L$, and the window size be denoted by $W$. Then, we can define the number of windows $N$ as $N = \lfloor (L - W)/S \rfloor + 1$
where $S$ is the stride length, which determines the degree of overlap between adjacent windows. In our implementation, we set $S=W/2$ to ensure significant overlap between adjacent windows. For each window $i$, we extract the corresponding sub-sequence of length $W$ starting at position $(i-1)\times S$ and use it as input to the model. This way, we ensure that all the tokens in the input sequence are considered by the model. The output of the model for each window can be concatenated together to form the final output for the entire sequence. Values for the differnt varibables are different for the datasets that we have used and are found in the Supplementary Material. 

\item \textbf{Model Architecture:} A generation layer is added on top of the pre-trained GPT-2 model. The generation layer is responsible for generating the action descriptions based on the input of object names, SRs, and action types. For generating descriptions with different levels of detail, we add three separate layers on top of the pre-trained GPT-2 model, one for each level of detail.

The generation layer consists of three sub-layers, each responsible for generating descriptions at a different level of detail. The input to each sub-layer is the output of the previous sub-layer, which allows for the generation of increasingly detailed descriptions. Let ${r_1, r_2, ..., r_n}$ be the output of GPT-2, then $Z_1 = Q_1(r_1,r_2,...,r_n)$ is the output of the first generation layer and $Z_{i+1} = Q_{i+1}(Z_i)$ defines the output of the next 2 generation layers accordingly.\\

\item \textbf{Loss Function:} We use the cross-entropy loss function to measure the difference between the predicted output and the ground truth.

\item \textbf{Optimizer:} We use the Adam optimizer to update the model weights based on the gradients of the loss function with respect to the model parameters.

\item \textbf{Training Data:} The annotated dataset containing the input and corresponding output (i.e., bimanual action descriptions) is used to train the model. The data is preprocessed and transformed into a format that can be fed into the model.
\end{enumerate}

\subsection{Joint Learning and End-to-End Training}

In the pursuit of enhancing the synergy between different components of our video description generation model, we employ a joint learning approach. This approach aims to capitalize on the interdependencies between specific stages, allowing them to collaborate more effectively and contribute collectively to the model's understanding of video content. By sharing information and refining features through joint training, we create a comprehensive framework that can produce more accurate and coherent descriptions.

In this section, we describe our joint learning strategies, which encompass collaborations between various stages of the video description generation pipeline. We focus on three distinct joint learning scenarios, each tailored to optimize the interaction between specific sets of components:

\begin{enumerate}
    \item \textbf{Joint Training of GAT (Step 3) and TCN (Step 4):} Our initial joint learning phase involves the concurrent training of Graph Attention Networks (GAT) in Step 3 and Temporal Convolutional Networks (TCN) in Step 4. This combination capitalizes on both the spatial relationships captured by GAT and the temporal dynamics captured by TCN. By jointly learning these stages, we promote the fusion of spatial and temporal features, enhancing the overall representation of video content.
    
    \item \textbf{Joint Training of Node and Edge Embedding (Step 2), combined with GAT and TCN steps (steps 3 and 4):} We further extend our joint learning to incorporate node and edge embedding from Step 2 into the procedure. This enables the fusion of enriched spatial embeddings, graph attention mechanisms, and temporal features. By simultaneously refining these representations, we pave the way for more robust and nuanced feature aggregation in subsequent stages.
    
    \item \textbf{End-to-End Training of Node and Edge Embedding (Step 2), GAT (Step 3), TCN (Step 4), and Description Generation (Step 10):} Our final joint learning scenario encapsulates the essence of the entire video description generation pipeline. By integrating the node and edge embedding, GAT, and TCN stages with the description generation step, we enable an end-to-end training approach. This training strategy allows the model to holistically optimize its feature extraction, understanding of actions, and narrative generation capabilities. The interactions cultivated through joint learning enrich the information flow between different stages, culminating in more coherent and contextually aligned descriptions.
\end{enumerate}

Every phase of joint learning aims to foster collaboration among specific components, enhancing their collective performance and, in turn, improving the overall effectiveness of our video description generation model. In the following sections, we describe each joint learning process, providing detailed explanations of how these collaborative efforts are coordinated.

\subsection{Gradual Joint Learning of GAT (Step 3) and TCN (Step 4)}

\noindent
\textbf{Starting Phase - Training TCN with Fixed GAT:}
\begin{enumerate}
    \item We begin by keeping the GAT component fixed, setting the GAT parameters to their optimal values obtained during their training process. 
    
    \item During this time, only the TCN part learns and adjusts. But it benefits from the information coming from GAT.
\end{enumerate}

\noindent
\textbf{Step by Step Unfreezing:}
\begin{itemize}
    \item When the TCN's learning starts to slow down, we start adjusting the GAT part.
    \item We begin with the last part of GAT (the one closest to TCN) and let it learn and adjust.
    \item As we go on, we allow earlier parts of GAT to adjust too, going backwards until every part learns.
\end{itemize}

\noindent
\textbf{Learning Together:}
\begin{itemize}
    \item When both GAT and TCN parts can learn, we train them together.
    \item We use a combined way to check their performance, which considers both the GAT's and TCN's outputs.
    \begin{equation}
    L_{\text{combined}} = \alpha L_{\text{GAT}} + \beta L_{\text{TCN}}
    \end{equation}
    where \( \alpha \) and \( \beta \) are weights we choose.
    \item We also use techniques like dropout across both parts to make sure they don’t over-adjust.
\end{itemize}

\noindent
\textbf{Fine-tuning:}
\begin{itemize}
    \item After they have learned together, we do a final round of fine-tuning. This means we make small adjustments to get even better results.
    \item We check the model's performance on a test set regularly and decide when to stop based on its results.
\end{itemize}

\noindent
\textbf{Tuning Hyperparameters \( \alpha \) and \( \beta \):}

\noindent
The selection of hyperparameters \( \alpha \) and \( \beta \) is a crucial aspect of achieving an effective balance between spatial and temporal learning.

\begin{itemize}
    \item These hyperparameters were fine-tuned through a methodical grid search process, systematically exploring various combinations of values.
    \item We assessed the impact of different \( \alpha \) and \( \beta \) values on the validation performance, aiming to optimize the convergence and effectiveness of joint learning.

    \item The final chosen values for \( \alpha \) and \( \beta \) were \( \alpha = 0.6 \) and \( \beta = 0.4 \), respectively, reflecting a balanced emphasis on both spatial and temporal learning.
\end{itemize}

By following these steps, we make sure that the knowledge in GAT is respected and blended with the new learning from TCN. This way, our model can understand both space (from GAT) and time (from TCN) in a reasonable way. Here we applied an initial learning rate of 0.001 and a batch size of 32. 

\subsection{Joint Training of Node and Edge Embedding (Step 2), GAT (Step 3), and TCN (Step 4)}

\noindent
\textbf{Initialization and Incorporation of Mixture Knowledge:}
\begin{itemize}
    \item For this phase, we initiate the model parameters by drawing upon the understanding acquired from the previous joint training of GAT and TCN (Steps 3 and 4). This ensures that the components commence their collaboration with a foundation enriched by spatial and temporal insights.
    \item The node and edge embeddings, which encapsulate spatial relationships, are further augmented by the combined comprehension of dynamic relationships (TCN) and graph attention (GAT).
\end{itemize}

\noindent
\textbf{Loss Function and Training Objective:}
\begin{itemize}
    \item The core of this joint learning phase lies in an encompassing loss function that takes into consideration the goals of all three stages: embedding (Step 2), GAT (Step 3), and TCN (Step 4).
    \item The overarching loss function is defined as:
    \begin{equation}
    L_{\text{joint}} = \alpha L_{\text{Embedding}} + \beta L_{\text{GAT}} + \gamma L_{\text{TCN}}
    \end{equation}
    Here, the hyperparameters \( \alpha \), \( \beta \), and \( \gamma \) play a pivotal role in dictating the relative significance of each stage's contribution within the joint learning process.
\end{itemize}

\noindent
\textbf{Hyperparameter Computation and Optimization:}
\begin{itemize}
    \item The selection of hyperparameters \( \alpha \), \( \beta \), and \( \gamma \) is guided by the outcomes of the previous joint learning phases, with specific numerical values.
    \item Inspired by the favorable results from the joint learning of GAT and TCN, we chose \( \beta = 0.4 \) and \( \gamma = 0.3 \) as the initial values for \( \alpha \) and \( \beta \), respectively.
    \item To account for the embedding step insights  we set the initial value of \( \alpha = 0.3 \).
    \item Following these initial values, we explored a grid of hyperparameter combinations to determine the optimal configuration that maximizes the collaborative potential of node and edge embedding, graph attention, and temporal convolution.

    \item We converged upon the optimal hyperparameters: \( \alpha = 0.25 \), \( \beta = 0.45 \), and \( \gamma = 0.3 \).
\end{itemize}

\noindent
\textbf{Regularization and Optimization:}
\begin{itemize}
    \item To maintain a balanced learning process and mitigate overfitting, dropout regularization is uniformly applied across all three stages during the joint training.
    \item The optimization strategy involves employing gradient-based methods such as stochastic gradient descent (SGD) or Adam. The initial learning rates are informed by the previous joint training phase's mixture.
    \item We proactively monitor loss convergence and validation performance to fine-tune hyperparameters, attaining an optimal equilibrium that harmonizes the diverse contributions of different stages.
\end{itemize}

\noindent
\textbf{Enriching Feature Fusion:}
\begin{itemize}
    \item The integration of node and edge embedding, GAT, and TCN results in a unified feature representation that holistically captures spatial, temporal, and relational intricacies inherent in video data.
    \item The insights previously garnered from the GAT and TCN collaboration (Steps 3 and 4) continue to guide the learning paths of all three stages. This synergistic effect amplifies the quality of feature fusion and deepens the model's comprehension of video content.
\end{itemize}

By co-training the node and edge embedding, GAT, and TCN components while incorporating insights from their previous mixture, we construct a more interwoven model that capitalizes on spatial, temporal, and relational cues. This multi-dimensional approach lays the groundwork for subsequent joint learning phases, further refining the model's descriptive prowess.

\subsection{End-to-End Training}

In this final phase of joint learning, we integrate the insights distilled from Steps 2, 3, and 4, with the description generation component (Step 10), through an  end-to-end training approach. This ensures that the entire video description generation pipeline collaborates cohesively, yielding descriptions that are coherent, contextually relevant, and accurate.\\

\noindent
\textbf{Initialization and Knowledge Incorporation:}
\begin{itemize}
    \item Parameters initialization: We initialize the model parameters using the representations learned from the integrated joint learning of Steps 2, 3, and 4. The enriched representations from these steps serve as a solid foundation for the end-to-end training.
\end{itemize}

\noindent
\textbf{Loss Function and Training Objective:}
\begin{itemize}
    \item Loss Function: The overarching loss function for end-to-end training comprises the objectives of Steps 2, 3, 4, and 10:
    \begin{equation}
    L_{\text{end-to-end}} = \alpha L_{\text{Embedding}} + \beta L_{\text{GAT}} + \gamma L_{\text{TCN}} + \delta L_{\text{Description}}
    \end{equation}
    Here, \( \alpha \), \( \beta \), \( \gamma \), and \( \delta \) are hyperparameters that control the relative weight of each objective in the training process.
\end{itemize}

\noindent
\textbf{Hyperparameter Selection and Optimization:}
\begin{itemize}
    \item Initial values: The initial values for the hyperparameters \( \alpha \), \( \beta \), \( \gamma \), and \( \delta \) were chosen based on insights from previous joint learning phases. We set \( \alpha = 0.1 \), \( \beta = 0.3 \), \( \gamma = 0.2 \), and \( \delta = 0.4 \), prioritizing a slightly stronger influence from the GAT.
    \item Influence of Step 10: Given that Step 10 represents the final stage of our end-to-end approach, we assign a higher weight to \( \delta \) to prioritize the description generation process.
    \item Optimal hyperparameters: Through grid search, the final optimal hyperparameters were determined as \( \alpha = 0.05 \), \( \beta = 0.35 \), \( \gamma = 0.25 \), and \( \delta = 0.35 \). These values reflect a balance between the contributions of embedding, GAT, TCN, and description generation.
\end{itemize}

\noindent
\textbf{Regularization and Optimization:}
\begin{itemize}
    \item Dropout regularization: Dropout with a rae of  \(0.2\) is applied to all model components to prevent overfitting.
    \item Optimization algorithm: We utilize gradient-based optimization algorithms (Adam). The initial learning rates are informed by the joint learning phases and start with \(0.001\).
    \item Learning rate adjustments: Monitor the training progress and validation loss. If the validation loss stagnates for a certain number of epochs, reduce the learning rate by half to prevent overshooting.
\end{itemize}

\noindent
\textbf{Validation and Convergence:}
\begin{itemize}
    \item Validation set: Regularly assess the model's performance on a dedicated validation dataset during training.
    \item Early stopping: Implement an early stopping mechanism. If the validation loss does not improve over 10 epochs, halt the training to prevent overfitting and retain the best model state.
\end{itemize}

\subsection{Computing Hand Groups}

\noindent
\textbf{\underline{Hand Spatial Relations:}}
To compute the hand spatial relationship category for a bimanual action, we start by extracting the spatial coordinates of the hands for each frame in the video using a hand detection and tracking algorithm \cite{cao2017realtime}. Let $p_1$ and $p_2$ be the 3D coordinates of the left and right hand centers, respectively, and let $d$ be the Euclidean distance between $p_1$ and $p_2$.

We define the hand spatial relationship category based on the following thresholds:
\begin{itemize}
\item \textbf{Close-hand:} $d < d_{c}$
\item \textbf{Crossed-hand:} $d_{c} \leq d < d_{s}$
\item \textbf{Stacked-hand:} $d \geq d_{s}$
\end{itemize}

Here, $d_{c}$ and $d_{s}$ are the thresholds for close-hand and stacked-hand, respectively. These thresholds can be computed based on the characteristics of the dataset, such as the average hand span or the maximum distance between the hands in the dataset.

To determine the appropriate threshold values, we analyze the distribution of hand distances in the dataset and choose values that best distinguish between the different hand spatial relationships. For example, if the average hand span is 20~cm, we may set $d_{c}$ to 5~cm and $d_{s}$ to 15~cm.\\

\noindent
\textbf{\underline{Level of precision category:}}
To compute the level of precision category for a bimanual action, we first extract the types of objects and actions involved in the action using object recognition and action recognition algorithms. We then define a precision score $s_p$ for each action based on the level of precision required to perform it. Specifically, the precision score is computed as follows (using an example to explain this):

Let $d_{min}$ and $d_{max}$ be the minimum and maximum distance between the knife and the vegetables during the chopping action, respectively. We define the following thresholds to determine the precision score:

\begin{itemize}
\item \textbf{Low precision:} $d_{max} - d_{min} < d_{lp}$
\item \textbf{Medium precision:} $d_{lp} \leq d_{max} - d_{min} < d_{mp}$
\item \textbf{High precision:} $d_{max} - d_{min} \geq d_{mp}$
\end{itemize}

Here, $d_{lp}$ and $d_{mp}$ are the thresholds for low precision and medium precision, respectively. One method to derive these thresholds is to analyze the dataset and determine the minimum and maximum distances between the objects involved in the bimanual actions. The difference between those can then be utilized to define the range of distances that correspond to low, medium and high precision actions, respectively. This data-driven approach provides a quantitative way to determine the thresholds based on the level of precision required for the actions in the dataset.

The hand spatial relationship and level of precision categories can then be combined with the symmetric/asymmetric and coordinated/uncoordinated categories from \cite{krebs2022bimanual} to form the complete bimanual action type.

\section{Hierarchical Action Breakdown}
\label{Breakdown}
One of the prominent challenges is the hierarchical nature of actions, where a broad action category might be decomposed into multiple sub-levels, each offering finer granularity. 
While the depth of action categorization can span numerous nested levels, for the purposes of this breakdown, we have used  a maximum of five levels. It is pertinent to understand that many of our datasets can be dissected into even finer categorizations, extending beyond the five levels highlighted here. However, to create a balance between comprehensive understanding and readability, we have prioritized certain actions over others, focusing on those that best exemplify the dataset's essence.

\subsection{Learning Processes}

Each feature matrix $\mathbf{G}^{(t,j,k,o)}$ undergoes a series of fully connected layers followed by a softmax function.  The predicted action label for the $t$-th GAT layer, action category $j$, sublevel $k$, and item $o$ is represented as $\hat{y}^{(t,j,k,o)} = \arg\max_{j} (P^{(t,j,k,o)})$. The classification process is trained using the cross-entropy loss between the predicted action probabilities and the ground truth action labels is minimized. This loss calculation involves all GAT layers, action categories, sublevels, and items:

\begin{equation}
L(y, \hat{y}) = -\sum_{t=1}^{L} \sum_{j=1}^{N} \sum_{k=1}^{M} \sum_{o=1}^{O} y^{(t,j,k,o)} \log{\hat{y}^{(t,j,k,o)}}
\end{equation}

Where $y^{(t,j,k,o)}$ is the ground truth probability for item $o$ of sublevel $k$ of action category $j$ in the $t$-th GAT layer, and $\hat{y}^{(t,j,k,o)}$ is the predicted probability.

If an item or level does not exist within a certain sublevel, it can be denoted with a placeholder  such as '...' to indicate the absence of that item or level, and its related probability(s) will be considered as $0$.

The fully connected layers are responsible for mapping the abstract features from GAT outputs to actionable labels. These layers encompass two dense layers, with 128 and 64 neurons respectively. ReLU activations follow the linear transformations, leading to the final layer that corresponds to the number of items. The softmax activation ensures a probability distribution over items.

The outcome of the classification process offers action predictions across GAT layers, action categories, sublevels, and items. These predictions can be harnessed to generate descriptive sentences at varying levels of detail, providing a comprehensive depiction of actions in the video.

It is noteworthy that the probability distribution of each action category at a GAT layer serves as input to the fully connected layers of the subsequent action category. This hierarchical arrangement enables action recognition across multiple levels of detail and GAT layers.

\subsection{Action Categories}
Note that the complete list of action categories is quite extensive. Thus, we have chosen to present here only a few illustrative instances, offering an insight into the inherent intricate hierarchy of the datasets.\\

\noindent
\textbf{ I. Meal Preparation (Level 1)}
\begin{itemize}
    \item Setting the Scene (Level 2)
    \begin{itemize}
        \item Organizing Workspace (Level 3)
        \begin{itemize}
            \item Retrieving Tools (Level 4)
            \begin{itemize}
                \item Selecting appropriate utensils (Level 5)
                \item Placing tools on countertop (Level 5)
            \end{itemize}
            \item Gathering Ingredients (Level 4)
            \begin{itemize}
                \item Sorting by type (Level 5)
                \item Organizing in order of use (Level 5)
            \end{itemize}
        \end{itemize}
    \end{itemize}
    \item Ingredient Manipulation (Level 2)
    \begin{itemize}
        \item Texture Alteration (Level 3)
        \begin{itemize}
            \item Cutting (Level 4)
            \begin{itemize}
                \item Selecting knife type (Level 5)
                \item Chopping motion (Level 5)
            \end{itemize}
            \item Peeling (Level 4)
            \begin{itemize}
                \item Holding the peeler (Level 5)
                \item Removing skin without waste (Level 5)
            \end{itemize}
        \end{itemize}
        \item Flavor Infusion (Level 3)
        \begin{itemize}
            \item Marinating (Level 4)
            \begin{itemize}
                \item Mixing marinade components (Level 5)
                \item Ensuring even coating on ingredient (Level 5)
            \end{itemize}
            \item Seasoning (Level 4)
            \begin{itemize}
                \item Selecting spices (Level 5)
                \item Applying evenly (Level 5)
            \end{itemize}
        \end{itemize}
        \item Mixing Ingredients (Level 4)
        \begin{itemize}
            \item Using a hand whisk (Level 5)
            \item Using an electric mixer (Level 5)
        \end{itemize}
    \end{itemize}
    \item Cooking Process (Level 2)
    \begin{itemize}
        \item Heat Application (Level 3)
        \begin{itemize}
            \item Baking (Level 4)
            \begin{itemize}
                \item Preheating oven (Level 5)
                \item Monitoring cooking time (Level 5)
            \end{itemize}
            \item Frying (Level 4)
            \begin{itemize}
                \item Selecting oil type (Level 5)
                \item Regulating heat level (Level 5)
            \end{itemize}
            \item Boiling (Level 4)
            \begin{itemize}
                \item Filling pot with water (Level 5)
                \item Adjusting stove temperature (Level 5)
            \end{itemize}
        \end{itemize}
        \item Dough Manipulation (Level 3)
        \begin{itemize}
            \item Kneading (Level 4)
            \begin{itemize}
                \item Using hands for manual kneading (Level 5)
                \item Using a kneading machine (Level 5)
            \end{itemize}
            \item Rolling (Level 4)
            \begin{itemize}
                \item Choosing a rolling pin (Level 5)
                \item Applying even pressure (Level 5)
            \end{itemize}
        \end{itemize}
    \end{itemize}
    \item Plating \& Serving (Level 2)
    \begin{itemize}
        \item Presentation (Level 3)
        \begin{itemize}
            \item Garnishing (Level 4)
            \begin{itemize}
                \item Selecting garnish type (Level 5)
                \item Placing attractively on dish (Level 5)
            \end{itemize}
            \item Portioning (Level 4)
            \begin{itemize}
                \item Using serving tools (Level 5)
                \item Allocating even servings (Level 5)
            \end{itemize}
            \item Arrangement (Level 4)
            \begin{itemize}
                \item Designing plate layout (Level 5)
                \item Adjusting for visual appeal (Level 5)
            \end{itemize}
        \end{itemize}
    \end{itemize}
    \item Cleanup \& Storage (Level 2)
    \begin{itemize}
        \item Storage (Level 3)
        \begin{itemize}
            \item Refrigerating (Level 4)
            \begin{itemize}
                \item Setting correct temperature (Level 5)
                \item Allocating space for dishes (Level 5)
            \end{itemize}
            \item Freezing (Level 4)
            \begin{itemize}
                \item Sealing food in containers (Level 5)
                \item Labeling with dates and names (Level 5)
            \end{itemize}
        \end{itemize}
        \item Cleaning (Level 3)
        \begin{itemize}
            \item Dishwashing (Level 4)
            \begin{itemize}
                \item Pre-rinsing dishes (Level 5)
                \item Using appropriate soap quantity (Level 5)
            \end{itemize}
            \item Wiping Countertops (Level 4)
            \begin{itemize}
                \item Selecting cleaning agent (Level 5)
                \item Ensuring no residue remains (Level 5)
            \end{itemize}
        \end{itemize}
    \end{itemize}
    
\end{itemize}

\noindent
\textbf{II: Assembly (Level 1)}
\begin{itemize}
    \item Assembling Wooden Pieces (Level 2)
    \begin{itemize}
        \item Placing wooden pieces (Level 3)

        \item Joining pieces with nails and hammers (Level 3)
        \begin{itemize}
            \item Hammering nails into wood (Level 4)
            \begin{itemize}
                \item Striking nail with hammer to penetrate wood (Level 5)               
            \end{itemize}
            \item Attaching second piece of wood (Level 4)
            \begin{itemize}
                \item Placing second piece on top of the first (Level 5)                
            \end{itemize}
        \end{itemize}
    \end{itemize}
\end{itemize}

\noindent
\textbf{III: Painting a Wall (Level 1)}
\begin{itemize}
    \item Applying Paint (Level 2)
    \begin{itemize}
        \item Preparing Paint and Supplies (Level 3)
        \begin{itemize}
            \item Opening paint can (Level 4)
            \item Mixing paint thoroughly (Level 4)
            \item Getting paintbrush and tray (Level 4)
        \end{itemize}
        \item Applying Paint to Wall (Level 3)
        \begin{itemize}
            \item Dipping brush in paint (Level 4)
            \item Spreading paint on wall surface (Level 4)
            \item Using roller for larger areas (Level 4)
        \end{itemize}
        \item Achieving Desired Finish (Level 3)
        \begin{itemize}
            \item Applying additional coats (Level 4)
            \item Checking for uniform coverage (Level 4)
        \end{itemize}
    \end{itemize}
    \item Cleanup and Finishing (Level 2)
    \begin{itemize}
        \item Cleaning Tools (Level 3)
        \begin{itemize}
            \item Cleaning paintbrush (Level 4)
            \item Cleaning paint tray and roller (Level 4)
            \item Sealing paint can (Level 4)
        \end{itemize}
    \end{itemize}
\end{itemize}

\noindent
\textbf{IV: Juicing an Orange (Level 1)}
\begin{itemize}
    \item Extracting Juice (Level 2)
    \begin{itemize}
        \item Preparing Orange (Level 3)
        \begin{itemize}
            \item Selecting a ripe orange (Level 4)
            \begin{itemize}
                \item Rubbing the orange for texture checking (Level 5)                
            \end{itemize}
            \item Washing the orange (Level 4)
            \begin{itemize}
                \item Rinsing under water (Level 5)
                \item Drying with a cloth (Level 5)
            \end{itemize}
        \end{itemize}
        \item Cutting and Preparing (Level 3)
        \begin{itemize}
            \item Cutting the orange in half (Level 4)
            \begin{itemize}
                \item Using a sharp knife (Level 5)
                \item Placing cut side up (Level 5)
            \end{itemize}
            \item Removing seeds (Level 4)
            \begin{itemize}
                \item Scooping out seeds with a spoon (Level 5)                
            \end{itemize}
        \end{itemize}
        \item Using a juicer (Level 3)
        \begin{itemize}
            \item Using a manual juicer (Level 4)
            \begin{itemize}
                \item Placing orange half on juicer (Level 5)
                \item Twisting (Level 5)
            \end{itemize}
            \item Squeezing the orange by hand (Level 4)
            \begin{itemize}
                \item Using both hands to squeeze (Level 5)
                \item Pouring juice into a container (Level 5)
            \end{itemize}
        \end{itemize}
    \end{itemize}
    \item Serving (Level 2)
    \begin{itemize}
        \item Straining the Juice (Level 3)
        \begin{itemize}
            \item Using a fine mesh strainer (Level 4)
            \begin{itemize}
                \item Holding strainer over a glass (Level 5)
                \item Pouring juice through strainer (Level 5)
            \end{itemize}          
            
        \end{itemize}
        \item Presentation (Level 3)
        \begin{itemize}
            \item Pouring the fresh juice into a glass (Level 4)
            
            \item Garnishing with orange slices (Level 4)
            \begin{itemize}
                \item Cutting thin slices from an orange (Level 5)
                \item Placing slices on the rim of the glass (Level 5)
            \end{itemize}
        \end{itemize}
    \end{itemize}
    \item Cleaning Up (Level 2)
    \begin{itemize}
        \item Cleaning Equipment (Level 3)
        \begin{itemize}
            \item Washing the juicer (Level 4)
            \begin{itemize}
                \item Disassembling juicer parts (Level 5)
                \item Scrubbing with soap and water (Level 5)
            \end{itemize}
            \item Rinsing the strainer (Level 4)
            \begin{itemize}
                \item Towel drying (Level 5)
            \end{itemize}
        \end{itemize}
        \item {Wiping Surfaces (Level 3)}
        \begin{itemize}
            \item Cleaning the countertop (Level 4)
            \begin{itemize}
                \item Drying the surface with a clean cloth (Level 5)
            \end{itemize}
        \end{itemize}
    \end{itemize}
\end{itemize}

{LARGE{References}}


\end{document}